\RequirePackage{fix-cm}
\documentclass[twocolumn]{svjour3}          
\smartqed  
\usepackage{graphicx}
\usepackage{graphicx}
\usepackage{subfigure}  
\usepackage{array}
\usepackage{longtable}
\begin{document}

\title{Knowledge Representation in Learning Classifier Systems: A Survey
}


\author{Farzaneh Shoeleh \and Mahshid Majd \and Ali Hamzeh \and Sattar Hashemi}


\institute{F. Shoeleh \at
              Computer Science and Engineering Dept. Shiraz University, Shiraz, Iran. \\
              \email{shoeleh@cse.shirazu.ac.ir}           
           \and
           M. Majd \at
              \email{majd@cse.shirazu.ac.ir}
           \and
           A. Hamzeh \at
              \email{ali@cse.shirazu.ac.ir}
           \and
           S. Hashemi \at
              \email{s\_hashemi@shirazu.ac.ir}
}

\date{Received: date / Accepted: date}

\maketitle

\begin{abstract}
Knowledge representation is a key component to the success of all rule based systems including learning classifier systems (LCSs). This component brings insight into how to partition the problem space what in turn seeks prominent role in generalization capacity of the system as a whole. Recently, knowledge representation component has received great deal of attention within data mining communities due to its impacts on rule based systems in terms of efficiency and efficacy. The current work is an attempt to find a comprehensive and yet elaborate view into the existing knowledge representation techniques in LCS domain in general and XCS in specific.   

To achieve the objectives, knowledge representation techniques are grouped into different categories based on the classification approach in which they are incorporated. In each category, the underlying rule representation schema and the format of classifier condition to support the corresponding representation are presented. Furthermore, a precise explanation on the way that each technique partitions the problem space along with the extensive experimental results is provided.  To have an elaborated view on the functionality of each technique, a comparative analysis of existing techniques on some conventional problems is provided. We expect this survey to be of interest to the LCS researchers and practitioners since it provides a guideline for choosing a proper knowledge representation technique for a given problem and also opens up new streams of research on this topic.

\keywords{Learning Classifier Systems \and XCS \and Knowledge Representation \and Rule Based Systems}
\end{abstract}

\section{Introduction}
\label{intro}
The first framework of Learning Classifier System (LCS) labeled "cognitive system" was introduced more than 30 years ago by John H. Holland (Holand, 1976). LCSs were originally inspired by the general principles of Darwinian evolution and cognitive learning. The LCS framework was reformed to use reinforcement learning techniques such as Q-learning (Sutton and Barto, 1998) in order to ensure appropriate reward estimation and propagation. LCS is also known as rule-based evolutionary online learning system. It is a heuristic method in which a population of production systems are consisted and adapted by using genetic algorithm and reinforcement learning techniques. Each production system can cover small region of environment and represent some portions of the overall solution. Therefore, a LCS system is able to solve a problem by using the best evolved production systems in its population. 

According to the solution encoding methodology, LCS has been addressed from two different points of view: the Pittsburgh Classifier System (Smith, 1980), usually referring to the models which have been inspired by the work of Smith and De Jong at the University of Pittsburg (Smith, 1980; De Jong, 1988; Smith, 1983), and Michigan Classifiers Systems that are usually the models which have been inspired by Holland's work at the University of Michigan (Holand, 1976; Holand, 1995). In Pittsburg approach, each population member is a production system and GA selects the best one as a complete solution of a given problem. In Michigan approach, the population members are individual rules, and the whole population forms the solution of the given problem. A rewarding mechanism is needed to reward and penalize bad rules.
 
The research in learning classifier systems sprang up in the 1980s. Early classifier systems were known as modeling tools. But they lost this initial characterization with rise of the abilities in the area of machine learning, especially in classification problems and the interest in reinforcement learning and autonomous agents. The year 1995 marked as a milestone in LCS researches due to Wilson's flavor of Holland's recipe. Wilson revised the main structure of Holland's LCS by simplifying it and changing its learning mechanism to use reinforcement learning techniques introduced in (Sutton and Barto, 1998). Nowadays the most popular Michigan system is an evolution of zeroth level classifier system, ZCS (Wilson, 1994), called accuracy based learning classifier system, XCS (Wilson, 1995), which can be considered as a milestone in classifier system research. XCS was the first LCS wherein the fitness of each rule is defined based on its accuracy of the payoff prediction instead of the receiving reward itself. Also, it has another main feature which uses the action set to define the environment niches. In Lanzi's view, the effectiveness of XCS as a machine learning paradigm is that \textit{"XCS was the first classifier system to be both general enough to allow applications to several domains and simple enough to allow duplication of the presented results"} (Lanzi, 2008). 

From 1995 to now, the applicability of XCS has been extended to a wide range of applications including computational economics (Schulenburg and Ross, 1999; Wong and Schulenburg, 2007), classification and data mining (Bull et al., 2008), autonomous robotics (Patel and Dorigo, 1994; Studley, 2005; Studley and Bull, 2005), power distribution network (Vargas et al., 2004), traffic light control (Bull et al., 2004), function approximation tasks (Wilson, 2002) and many more (Bull, 2004; Lanzi et al., 2000; Bull et al., 2008). So, the current LCS researches are very diverse and pervasive and done better than ever.

There are many valuable researches to show the effectiveness and the power of LCS in many domains. Some of these researches provided a description of the overall advances of the LCS field as a survey study. For example, in (Lanzi, 2008), the author tried to answer this question\textit{ "What has happened to learning classifier systems in the last decade?"} and examined the current state of learning classifier systems research. Indeed, the aim of this study is to emphasize recent developments and the state-of-the-art Learning Classifier Systems. Bull et al. brought together the works on the use of LCS for data mining problems (Bull et al., 2008). LCSs have proved its efficiency at solving online and offline classification tasks. Bacardit et al. gave a summary of past, present and future LCS researches in (Bacardit et al., 2008). They tried to take a look back at the LCS realm and discuss which challenges and opportunities are laying ahead for successful system applications in various domains. As LCS can be seen as a learning system which is able to solve reinforcement learning problems, a study (Sigaud and Wilson, 2007) was done which focuses more on the sequential decision domains than on automatic classification. In (Urbanowicz and Moore, 2009), after introducing a description of basic LCS framework, a historical review of major advancements and a roadmap of algorithmic components are provided. It also emphasizes the differences between alternative LCS implementation and their problem domains. The main aim of such studies is to provide accessible and comprehensible principles and different backgrounds for researches interested in developing their own LCS to achieve a specific goal. 

To the best of our knowledge, most of the existing surveys have tried to investigate the general behavior of LCS in terms of its common issues, basic components and major applications. These studies can be considered as appropriate resources for who are interested in LCS research field providing that they give a general perspective of the current research opportunities and challenges in LCSs area. But they might not be efficient enough to show the progress of LCSs regarding to the improvement of their main components to handle new issues and challenges. In other words, according to the large number of studies dedicated to each of the main components of LCSs, surveys individually focusing on each one would be of great use provided that they show how such component can be modified to improve the performance of a LCS like XCS in solving a given problem. Consequently, this survey attempts to provide a detailed description of LCS concentrating on one of its major components known as knowledge representation. . Recently, knowledge representation component has received great deal of attention within data mining communities due to its impacts on rule based systems in terms of efficiency and efficacy. It is worth mentioning that it is the first step to identify the environmental properties and it supports an interaction between the environment and the learning system. The current work is an attempt to elaborate view into the existing knowledge representation in LCS domain in general and XCS in specific. The knowledge representation techniques are grouped into different categories based on the classification approach in which they are incorporated. In each category, the underlying rule representation schema, the format of classifier condition and a precise explanation on how it can partition the problem space along with the experimental results are provided. Additionally, to have an elaborated view on the functionality of each technique, a comparative analysis of existing techniques on some conventional problems is provided. It is worth mentioning that the main focus of current study is not explicitly on presenting an alternative knowledge representation scheme, rather it focuses on providing a guideline to choose a proper representation technique for a given problem which is nowadays human need. Undoubtedly, addressing the none-trivial recent issue is of great importance. We hope that this survey facilitates a better understanding of the different streams of research on this topic and provides a guideline for choosing a proper knowledge representation technique for a given problem. Furthermore, it can be of interest to the LCS researchers and practitioners and also opens up new streams of research on this topic. 

The rest of this paper is organized as follows; at first in Section 2, we briefly introduce learning classifier systems and provide a description of XCS in detail. Section 3 describes the knowledge representation component and the various works that have been done to enhance this component. We group knowledge representation techniques into different categories based on the classification approach in which they are incorporated.  In the subsections of Section 3, the underlying rule representation schema, the format of classifier condition and a precise explanation on how it can partition the problem space along with the experimental results are provided. In Section 4, to elucidate the functionalities of each technique a comparative analysis of existing techniques on some conventional problems is provided in general perspective. The last section includes summary and conclusion remarks.
\section{Learning Classifier Systems in Brief}
\label{sec:1}
Learning Classifier Systems belong to a family of machine learning techniques which combines genetic algorithm (GA) with the power of the reinforcement learning paradigm to solve a given problem. In Holland's recipe, as the inventor of LCS, it has four main components with specific goals: (1) a population of classifiers known as knowledge representation component, which represents the current system knowledge; (2) a performance component, which provides a proper input for system and an applicable output to apply to the learning environment; (3) a reinforcement (or credit assignment) component, which is responsible to distribute the incoming reward among the classifiers that are accountable for it; (4) a rule discovery component, which creates new rules through a covering mechanism and evolves the existing ones by using an evolutionary algorithm, usually a GA. Most of the developed models which had been proposed in the realm of Michigan style LCS were extended and improved the original idea of Holland's LCS, but kept these entire four main components. It is notable that the effectiveness of LCS to solve a problem depends on the proper interaction between these four components, therefore the compatibility between them must be considered in developing a new extension of LCS. 

In each learning step of a Michigan style LCS, the performance component perceives the environmental state (or input) through its detectors and performs the selected action through its effectors. Then, the environment eventually pays a reward to the system regarding the selected action's efficacy. Like other reinforcement learning techniques, LCS tries to suggest a solution which can maximize the amount of received reward through, a set of condition-action-prediction rules called classifiers which are produced and evolved to represent the current solution. On the other hand, the reinforcement component acts on the population to estimate the action values in each subproblem. The discovery component produces new rules for unseen subproblems and usually uses genetic algorithm to evolve the current solution. In the next section, a brief review on XCS (a milestone of LCS) and its main components is provided.

\subsection{XCS in brief}
\label{sec:2}
The year 1995 is marked a milestone in LCS researches due to the appearance of the most successful and popular Michigan style LCS named XCS. Recent analysis and researches (Wong and Schulenburg, 2007; Bull et al., 2008; Studley and Bull, 2005; Vargas et al., 2004) have shown that XCS has an especial architecture that makes it a competitive learning system in solving complex problems. Its architecture ingredients provide some promising properties such as its online learning capability, its noise robustness, the generality in the learning mechanism, and its continuous adaptation.

XCS contains a population of classifiers which is called [P]. This population can be empty in the beginning of the experiment, or be filled randomly. Each classifier in [P] is made up of different parts. These parts are: a condition usually from the alphabet \{0, 1, \#\}, an action which is usually an integer, and a set of associated parameters. These parameters are (1) a payoff prediction \textit{$P_j$}, which estimates the payoff that the system will receive when its action is applied to the corresponding environment; (2) the prediction error \textit{$\epsilon_j$}, (3) the fitness \textit{$F_j$}, and some other parameters such as \textit{exp}, \textit{num} and etc.

When XCS receives an environmental state, it forms the related match set [M]. This set includes those classifiers whose condition parts match the current environmental state. If no classifier matches, the covering operator will create a predefined number of classifiers which match the current input and insert them into the population and into [M]. If the covering operator causes the size of the population to grow over a predefined threshold \textit{N}, some other classifiers will be eliminated from the population regarding their fitness and experience (\textit{exp} parameter).

Then, for each action $a_k$, which is proposed by classifiers in $[M]$, the system computes a fitness weighted average $P_k$ using this equation: $P_k=\frac{\sum_{cl\in[M|a_k]} cl^{k}.F\times cl^{k}.P}{\sum_{cl\in[M]} cl^{k}.F}$ where $[M|a_k]$ is the subset of $[M]$ contains classifiers which propose the action $a_k$.This value is used as the bid of the corresponding action to win the current phase. Then, XCS chooses an action from those proposed in $[M]$ regarding its Explore/Exploit strategy. Finally, an action set $[A]$ is formed which consists of a subset of $[M]$ with the same action as the chosen one.

After that, the selected action is applied to the environment and a reward \textit{R} is received from the environment which may be zero. Then, the parameters of the involved classifiers are updated. In the sequential environments, the update procedure occurs for classifiers in the previous action set which is called hereafter $[A]^{-1}$. It is also notable that all updates are done using a Q-Learning update regime. To do so, the payoff \textit{P} is calculated as follows: $P=r_{-1}+\gamma\max_{a}{p_a}$ where $r_{-1}$ is the previous environmental reward which is 0 or 1000 for classification problem, $\gamma$ is the discount factor and pa is the predicted payoff for the applied action in the current trial. Then, classifier predictions in $[A]^{-1}$ are updated as follows: $P_j = P_j+\beta (R-{P_j})$ where $\beta$ is the learning rate in the Widrow-Hoff update rule. Next, the prediction errors are re-estimated using this equation:$\epsilon_j =\epsilon_j+\beta(|R-P_j|)$. Then the accuracy for the corresponding classifier is calculated as follows: $k_j=0.1 {(\frac{\epsilon_j} {\epsilon_0})^{-\nu}}$ for $\epsilon_j>\epsilon_0$, else 1.0. The parameter $\epsilon_0$ is termed the error threshold and $\nu$ is a positive integer and both of them are initiated at the beginning of the experiments. Then, the relative accuracy of each classifier is calculated as: $k'_j=\frac{k_j} {\sum_j{k_j}}$. And at last, this relative accuracy is used to update the classifier fitness: $F_j=F_j+\beta(|k'_j-F_j|)$. These updated values are used in another important component of XCS: The discovery component. On a predefined period based on the parameter $\theta_{ga}$, a GA is applied to $[A]^{-1}$. As usual, applying the GA consists of three phases: the selection, the crossover and the mutation. In the selection phase, two classifiers are selected with the proportionate selection operator regarding their fitness. The crossover operator is applied to the two selected parents at the rate of $\chi$. Then, at the rate of $\mu$, each allele of the generated offspring is mutated. The resulting offspring are inserted into the population. If the size of the population grows over a predefined threshold \textit{N}, two classifiers will be deleted from the population considering their fitness.

To improve the generalization capability of XCS, another concept is utilized: the subsumption deletion. When a classifier is added to the population by GA, it is checked against all classifiers in the population to find a sufficiently experienced and accurate classifier which covers the newly inserted classifier. If the covering classifier is found, then the new classifier will be eliminated from the population and the numerosity parameter (\textit{num}) of the covering classifier will be added by one. This process is called GA subsumption. A same process also occurs in the action set creation phase. In this phase, all classifiers in $[A]$ are matched against accurate and sufficiently experienced classifiers in $[A]$. If a classifier is covered by another classifier, it will be eliminated from the population and the numerosity parameter of the covering classifier will be increased by one.

\begin{figure}
  \includegraphics[width=0.5\textwidth]{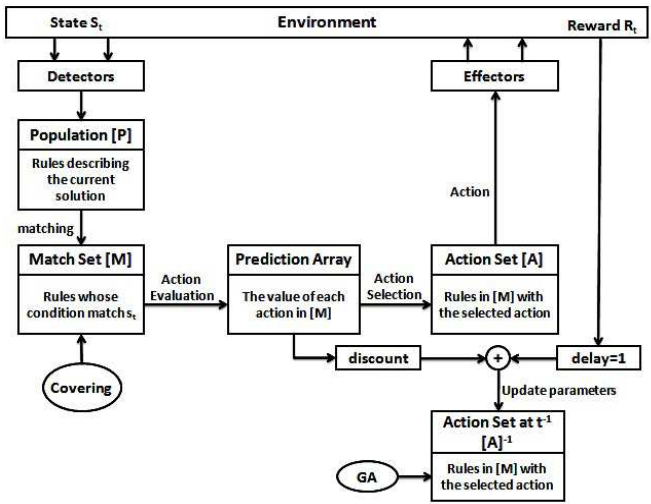}
\caption{Schematic illustration of XCS life cycle. In a typical iteration, XCS receives the current environmental state then forming $[M]$ from $[P]$. Next, the proper action is chosen to execute in the environment. The rule parameters would be updated according to the resulting reward. Finally, the GA may be applied on $[A]$.}
\label{fig:2}       
\end{figure}
Figure 1 gives an overall picture of XCS system, which is shown in interaction with an environment. Above XCS description is an abbreviate one, for more details; see the original XCS paper (Wilson, 1995) and (Butz and Wilson, 2001).

\section{Knowledge Representation}
\label{sec:3}
The first and main step to solve a problem is finding the proper realization of its definition and input space. In the other words, the solver system must be able to model the regularities of the problem space to make decision regarding its experiences. In rule based systems, their rule set is the one responsible for such modeling where each rule represents some portion of the problem space and makes the best decision in corresponding subspace. Consequently, a rule based system is able to solve a problem efficiently if the rule set can cover the whole problem space properly and also each rule makes effective decision. Therefore, it can be concluded that rule representation, i.e. knowledge representation, has an essential role in rule based systems. In particular, a rule based classifier like XCS can achieve all the three main objectives of classification task which are high accuracy, comprehensibility and compactness, provided that it has a proper rule representation. 

The first and main step to solve a problem is finding the proper realization of its definition and input space. In the other words, the solver system must be able to model the regularities of the problem space to make decision regarding its experiences. In rule based systems, their rule set is the one responsible for such modeling where each rule represents some portion of the problem space and makes the best decision in corresponding subspace. Consequently, a rule based system is able to solve a problem efficiently if the rule set can cover the whole problem space properly and also each rule makes effective decision. Therefore, it can be concluded that rule representation, i.e. knowledge representation, has an essential role in rule based systems. In particular, a rule based classifier like XCS can achieve all the three main objectives of classification task which are high accuracy, comprehensibility and compactness, provided that it has a proper rule representation. 

As mentioned earlier, knowledge representation is an essential component of XCS whose elements are certain number of classifiers. Each classifier has two main parts; a condition which represents the regularities of the problem regarding effective generalization, and, an action part that shows what decision being made in corresponding condition. These parts can be adapted for a particular purpose, without modifying the main structure of the system.

In the problems with binary inputs, the first and most commonly used syntax for classifier condition termed "Ternary" can be used as done in (Wilson, 1995; Holland and Reitman,1978; Schuurmans and Schaeffer, 1989). In this syntax, the condition part is simply represented by a fixed length bit string defined over the alphabet \{0, 1, \#\} where the \textit{'don't care'} symbol (\#) matches both one and zero. It is mentioned in (Schuurmans and Schaeffer, 1989) that \textit{"…Ternary representation can hardly model the relation among problem and there is an avoidable bias in system generalization capabilities…"}, it has three noticeable advantages; first, it is easy to analyze and the obtained rule set is well understood; second, it is proper for textual and categorical information; and third, any data are eventually changed into binary in a computer system. However, since most of real world data sets contain real valued, nominal or mixed attributed data, researchers encourage to proposed powerful representation which are applicable to the complex problems with such data sets.

Many representations have been developed to handle real valued data, such as, disjunctions of intervals (Wilson, 2000a; Wilson, 2000b; Stone and Bull, 2003; Dam et al., 2005a), ellipsoidal (Butz, 2005; Butz et al., 2006; Butz et al., 2008) and convex hulls (Lanzi and Wilson, 2006). Other general purpose representations have also been developed, such as, first order logic (Mellor, 2005; Mellor, 2006), fuzzy logic (Valenzuela-Rend$\acute{o}$n, 1991; Bonarini and Matteucci, 2007; Casillas et al., 2007; Orriols-Puig et al., 2008a), GP-like conditions (Lanzi and Perrucci, 1999; Lanzi, 2001; Lanzi, 2003; Wilson 2008; Preen and Bull, 2009), messy coding (Lanzi, 1999), tile coding (Lanzi et al. 2006) and neural network (Bull and O'Hara, 2002; Hurst and Bull, 2004; O'Hara and Bull, 2005; Howard et al., 2008; Dam et al., 2008) based representation. Table 1 shows the synopsis of different kinds of knowledge representation emerged in XCS. The next subsections arranged in such a way that they detail knowledge representation schemes provided in Table 1 respectively.  
\begin{center}
\begin{table*}
\caption{A summary of different rule representation methods developed for XCS and some of their features, such as; LCS algorithms/platforms, the year of proposing corresponding representation, and what articles have used such representation.}
\label{tab:1}       
\scriptsize
\begin{tabular}{lllll}
\hline\noalign{\smallskip}
{Rule Representation} & & {System} & {Introduced in}  & {Used in} \\
\noalign{\smallskip}\hline\noalign{\smallskip}
Interval Based   & Center Spread Representation & XCSR   & 
Wilson, 2000a & Wilson, 2000a; \\
Representation & Lower-Upper Bound Representation & XCSI & Wilson, 2000b & Wilson, 2001b; \\
& Unordered-Bound Representation & --- & Stone and Bull, 2003 & Wilson, 2001; \\
& Min-Percentage Representation & --- & Dam et al., 2005 & Bernad$\acute{o}$-Mansilla et al., 2001; \\
& & & & Wilson, 2002; \\
& & & & Hurst and Bull, 2004; \\
& & & & Gao et al., 2005; \\
& & & & Dam et al., 2005b;  \\
& & & & Hamzeh and Rahmani, 2005; \\
& & & & Gao et al., 2006;\\
& & & & Gao et al., 2007; \\
& & & & Hamzeh and Rahmani, 2007;\\
& & & & Gao et al., 2007;\\
& & & & Llor$\acute{a}$ et al., 2007;\\
& & & & Lanzi and Loiacono, 2007\\
\noalign{\smallskip}\hline
Ellipsoidal Based & Hyper sphere Representation & --- & 
Butz, 2005 & Butz et al., 2005; \\
Representation & Hyper ellipsoidal & --- & Butz, 2005 & Butz et al., 2006; \\
& General hyper ellipsoidal & --- & Butz, 2005 & Wilson, 2001;  Butz et al., 2008\\
\noalign{\smallskip}\hline

Convex Hull Based &  & --- & 
Lanzi and Wilson, 2006 & Lanzi and Wilson, 2006; \\
Representation & & & &\\
\noalign{\smallskip}\hline

Fuzzy Logic Based &  & Fuzzy LCS & 
Valenzuela-Rend$\acute{o}$n, 1991 & Valenzuela-Rend$\acute{o}$n, 1991 \\
Representation& &  ELF & Bonarini, 1993 & Bonarini, 1993; \\
& &  LFCS & Bonarini, 2000; & Bonarini, 1994; \\
& &  LFCS & Bonarini et al., 2000 &  Bonarini, 1998; \\
& &  FIXCS & Bonarini and Matteucci, 2007 &  Walter and Mohan, 2000;\\
& &  Fuzzy-XCS & Casillas  et al., 2004 &  Bonarini, 2000; \\
& &  Fuzzy-UCS & Orriols-Puig , 2007 & Bonarini et al., 2000; \\
& & & & Casillas  et al., 2004; \\
& & & & Casillas  et al., 2005; \\
& & & & Casillas  et al., 2007; \\
& & & & Orriols-Puig et al., 2007; \\
& & & & Bonarini and Matteucci, 2007; \\
& & & & Orriols-Puig et al., 2008a; \\
& & & & Orriols-Puig et al., 2008b\\
\noalign{\smallskip}\hline

First Order Logic Based &  & FOXCS & 
Mellor, 2005 & Mellor, 2005; \\
Representation& & & & Mellor, 2006; \\
& & & & Mellor, 2008; \\
\noalign{\smallskip}\hline

Messy Code &  & XCSm & 
Lanzi, 1999 & Lanzi, 1999; \\
Representation & & & & \\
\noalign{\smallskip}\hline

GP-like & S-expression & XCSL & 
Lanzi and Perrucci, 1999 & Lanzi and Perrucci, 1999; \\
Representation& Stack Based Representation  & --- & Lanzi, 2003&  Lanzi, 2001a; \\
& \textit{(PRN expression)} & & & Lanzi, 2001b; \\
& GEP & XCSF-GEP& Wilson 2008& Lanzi, 2003;;\\
& Dynamical Genetic Programming & DGP-XCS & Preen and Bull, 2009& Unold, 2005\\
&\textit{(graph based representation)} & & & Unold and Cielecki, 2005;\\
& & & &  Unold, 2007;\\
& & & &   Cielecki and Unold, 2007;\\
& & & &  Wilson,2008; \\
& & & & Preen and Bull, 2009 \\
\noalign{\smallskip}\hline

Neural Networks Based & Neural Network  & X-NCS & 
Bull and O'Hara, 2002 & Bull and O'Hara, 2002; \\
Representation& & NCS & Hurst and Bull, 2004 & Hurst and Bull, 2004;  \\
& & NLCS & Dam et al., 2008 & O'Hara and Bull, 2005;\\
& & XCSFNN & Loiacono and Lanzi, 2006 &   Loiacono and Lanzi, 2006; \\
& Fuzzy-Neural Network & X-NFCS & Bull and O'Hara, 2002 &  Dam et al., 2008\\
\noalign{\smallskip}\hline

Tile Coding Based&  & XCSF-RTC & 
Lanzi et al., 2006 & Lanzi et al., 2006; \\
Representation & & & & \\
\noalign{\smallskip}\hline

\end{tabular}
\end{table*}
\end{center}
Further on, we group the knowledge representation techniques into different categories described in each subsection based on the classification approach in which they are incorporated. In each category, we present the following issues: the underlying rule representation schema and the format of classifier condition to support the corresponding representation, a precise explanation bringing insight into how to partition the problem space and the extensive experimental results reported in the original papers.

\subsection{Interval Based Representation}

\label{sec:4}

The traditional ternary representation has been substituted with interval-based representation to manage continuous-valued inputs. It has been shown that this modified XCS can be effectively applied to real-valued problems (Bernad$\acute{o}$-Mansilla et al., 2001; Wilson, 2000b; Wyatt, 2004; Bull et al., 2008). In the interval-based representation condition part of each classifier is determined using intervals defined over each dimension. To define these intervals, four representation techniques have been introduced as follows:

\textbf{Center Spread Representation (CSR):} In (Wilson, 2000a), Wilson modified XCS and introduced CSR to handle real-valued inputs. In CSR, for the solution space of range $[p_{min},q_{max})$, an interval predicate $[p_i,q_i)_p$ is represented as a tuple $(c_i,s_i)_g$ where $c_i$ indicates the center of the interval and $s_i$ is a radius around $c_i$. Both $c_i$ and $s_i$ are real valued numbers. So, a classifier matches input $x$ if and only if $c_i-s_i<x_i<c_i+s_i$ for all dimension of $x$. 

Stone and Bull (Stone and Bull, 2003) analyzed the behavior of classifiers with this representation in XCS. They argued that using CSR causes the incomplete and many to one genotype to phenotype mapping. As an example, the possible values of i'th dimension vary between [0,10), the interval [7,10) in phenotype space can be mapped by (9,2),(8.5,1.5), (10,3), and many more.

Once nonlinear genotype to phenotype mapping has performed and mutation and crossover operators have applied, a truncation operator must be applied to classifiers to keep them in range. This operator introduces a bias in the distribution of intervals phonotype. Through this bias, the frequencies of $[p_{min},q_i)_p$ and $[p_i,q_{max})_p$ intervals increase as $q_i$ increases or $p_i$ decreases, respectively. Therefore, the $[p_{min},q_{max})_p$ interval would have much greater chance to be appeared in classifier conditions. 

\textbf{Lower-Upper Bound Representation or Min-Max Representation (MMR):} Wilson lately proposed another representation named MMR (Wilson, 2000b) for integer-valued inputs. It can also be used for real-valued inputs (Wilson, 2002). Here, an interval predicate $[p_i,q_i)_p$ is encoded as a tuple $[l_i,u_i)_g$ where $l_i=p_i$ and $u_i=q_i$. In real valued problems $l_i$,$u_i\in R$  and in integer-valued problems $l_i,u_i \in Z$. A classifier matches input $x$ if and only if $l_i<x_i<u_i$ for all dimension of $x$.

Stone and Bull (Stone and Bull, 2003) stated that this representation has complete and one to one genotype to phenotype mapping. Furthermore, this method overcomes the bias generated by truncation in CSR. In comparison with CSR, in MMR, the frequency of intervals of the form $[p_{min},q_i)_p$ and $[p_i,q_{max})_p$ is constant for all $p_i$ and $q_i$ and $[p_{min},q_{max})_p$ interval has same frequency as that of any other interval. But, another issue arises; the tuples have an ordering restriction $(l_i<u_i)$ and genetic operators might produce infeasible intervals.

\textbf{Unordered-Bound Representation (UBR):} In order to overcome the ordering issue of the previous method, Stone and Bull (Stone and Bull, 2003) proposed a new approach named unordered-bound representation (UBR) which is similar to MMR without ordering restriction. Thus, an interval predicate $[p_i,q_i)_p$ can be defined in the form of $[l_i,u_i)_g$ or $[l_i,u_i)_g$ where  $l_i\neq u_i$ and $l_i=p_i$ and $u_i=q_i$. In other words, the genotype to phenotype mapping is normally complete and two to one except where $q_i=p_i$ which is one to one. The mutation operator provides strong specialization pressure with UBR, unlike MMR and CSR, where the mutation operator yields a low level of specialization pressure.

Although UBR is able to obviate the problems of previous approaches, it raises new issues. The semantic of the genotypes is not permanent; that is, the gene presenting lower bound of interval in one generation may present the upper bound in the next generation. This issue is inconsistent with building block hypothesis which is the keystone in the architecture of GA and is known as one of the important necessities to run it (Goldberg, 1989).

\textbf{Min-Percentage Representation (MPR):} In (Dam et al., 2005a), Dam et al. purposed to overcome the problem of UBR by presenting new approach named Min-Percentage Representation which maintains the semantic of the genotype all over evolutionary run. In MPR, an interval predicate $[p_i,q_i)_p$ can be encoded as a tuple $[m_i,per_i)_g$ where $m_i=p_i$ and $per_i=\frac{q_i-p_i} {p_{max}-p_i}$. So a classifier matches input $x$ if and only if $m_i<x_i<m_i+per_i*(p_{max}-m_i)$ for all dimension of $x$.

In (Dam et al., 2005a), MRP and UBR were compared in 6-Real-Multiplexer and Checkerboard problems. Results indicated that the two techniques have equivalent performance and similar behavior in both problems. But, MRP is unlikely to change the semantic of the genotypes.
%

\begin{figure}[!htb]

\subfigure []{
\includegraphics[width=0.23\textwidth]{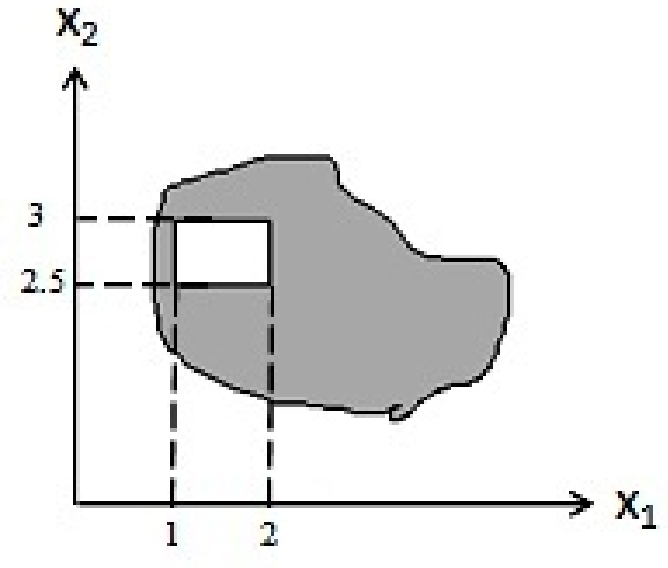}}
\subfigure []{
\includegraphics[width=0.23\textwidth]{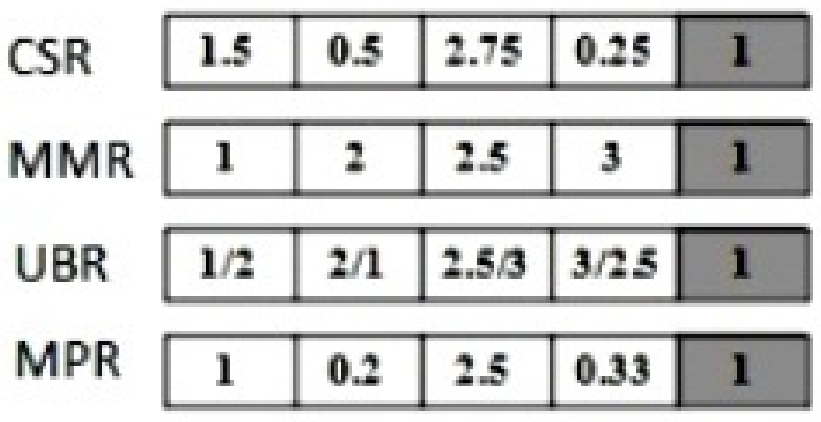}}
\caption{An example of interval based classifier in an arbitrary problem space. a) Visualization of the covered area by interval based classifier. b) The genotype of the classifier presented in (a) with four different types of interval based representation. The white cells identify the condition part and the gray cell identifies the action part of the corresponding classifier.}
\label{fig:2}       
\end{figure}

As illustrated in Figure 2, XCS with interval based representation uses hyper rectangles to partition the problem space. In (Butz et al., 2006) Butz showed, that the interval based condition makes difficulties to solve problems with oblique decision boundaries. He also showed that the learning takes longer time and the compactness of the model is not efficient when dealing with oblique boundaries. For example, XCSF (Wilson, 2002), one of the successful extensions of XCS, needs an enormous number of classifiers with hyper rectangle conditions to accurately approximate a nonlinear surface such as a circle. Interval based representation is well-suited in modeling axis-parallel boundaries, but in dealing with oblique boundaries, it is more advantageous to use a representation which partitions the problem space into complex regions.

\subsection{Ellipsoidal Based Representation}
\label{sec:5}
In (Butz, 2005), Butz suggested a new structure of classifier condition, based on defining hyper sphere and hyper ellipsoid shapes. There, three structures were proposed; first, the condition part of the classifiers represents a hyper sphere in the problem space. As hyper sphere has common radius in all dimensions, the condition part consists of a center and a deviation, that is, $C=\{\overrightarrow{m},\sigma\}=\{m_1,m_2,...,m_n,\sigma\}$. Second, the condition part represents an axis-parallel hyper ellipsoid which has different deviation in each dimension, that is, $C=\{\overrightarrow{m},\overrightarrow{\sigma}\}=\{m_1,m_2,...,m_n,\sigma_1,\sigma_2,...,\sigma_n\}$. Third, the condition part is redefined to present a general hyper ellipsoidal structure which is axis-independent unlike the previous one. So, it is defined by a center point in addition to elements of matrix $\Sigma$ named transformation matrix which indicates fully Mahalanobis distance metric of a hyper ellipsoid. This matrix shows stretch and rotation of the represented hyper ellipsoid. The condition part of a classifier is defined as follows: $C=\{\overrightarrow{m},\Sigma\}=\{m_1,m_2,...,m_n,\sigma_{1,1},\sigma_{1,2},...,\sigma_{(n,n-1)},\sigma_{(n,n)}\}$ where $\overrightarrow{m}$ shows the center of represented hyper ellipsoid and $\Sigma$ shows the transformation matrix of the condition. This structure is a general form of the first and second ones; hyper sphere is a hyper ellipsoid where the diagonal entries of $\Sigma$ are initialized with common value and all other entries are set to zero, and axis-parallel hyper ellipsoid has a diagonal matrix as $\Sigma$. 
In the general form, angular orientation and stretch of the represented hyper ellipsoid are implicitly encoded in $\Sigma$. Due to the redundancy of this encoding; the mutation and crossover operators may not act beneficially (Butz et al., 2008). This would decrease the reproductive opportunities of the successful classifiers and lead the evolutionary progress to slow down. To solve this problem, Butz et al. (Butz et al., 2006; Butz et al., 2008) investigated another condition representation which is able to explicitly codify the rotation of the desired hyper ellipsoid in its structure. So, the condition part of each classifier is expressed by three vectors, $C=\{\overrightarrow{m},\overrightarrow{\sigma},\overrightarrow{\gamma}\}$; a vector $\overrightarrow{m}$ that indicates the center point, a vector $\overrightarrow{\sigma}=(\sigma_1,\sigma_2,...,\sigma_n)^T$ which represent the stretch, and a vector \overrightarrow{\gamma} with size of $(_2^n)$ to point out the orientation angles of the corresponding hyper ellipsoid. Figure 3 highlights how a classifier with mentioned representations can cover a partition of the input space.
\begin{figure}[!htb]
\subfigure []{
\includegraphics[width=0.23\textwidth]{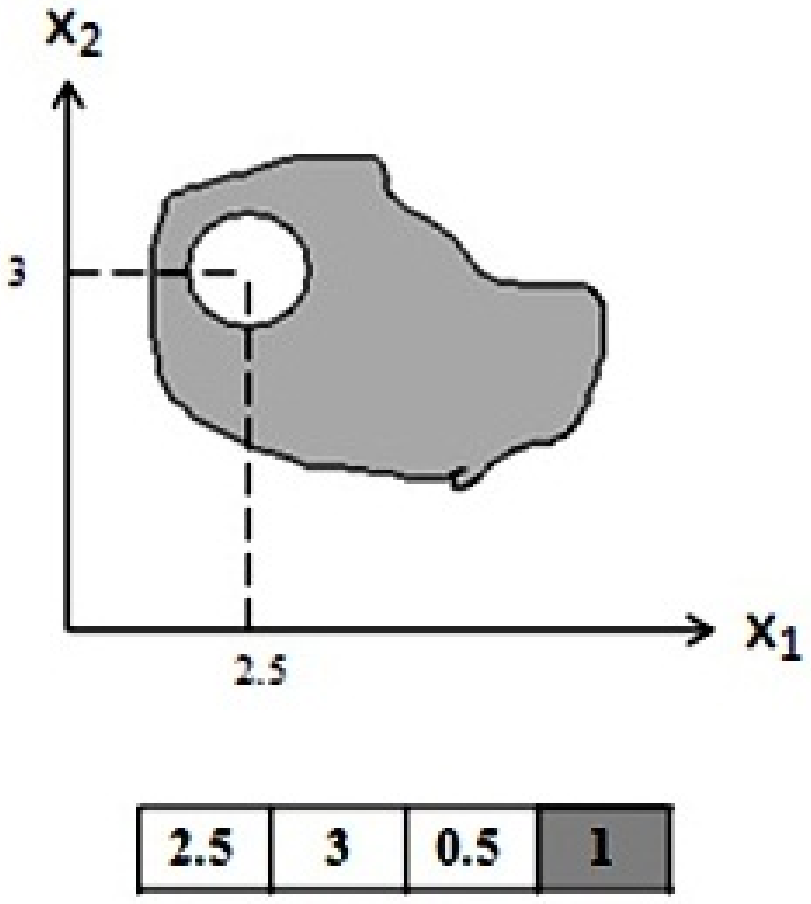}}
\subfigure []{
\includegraphics[width=0.23\textwidth]{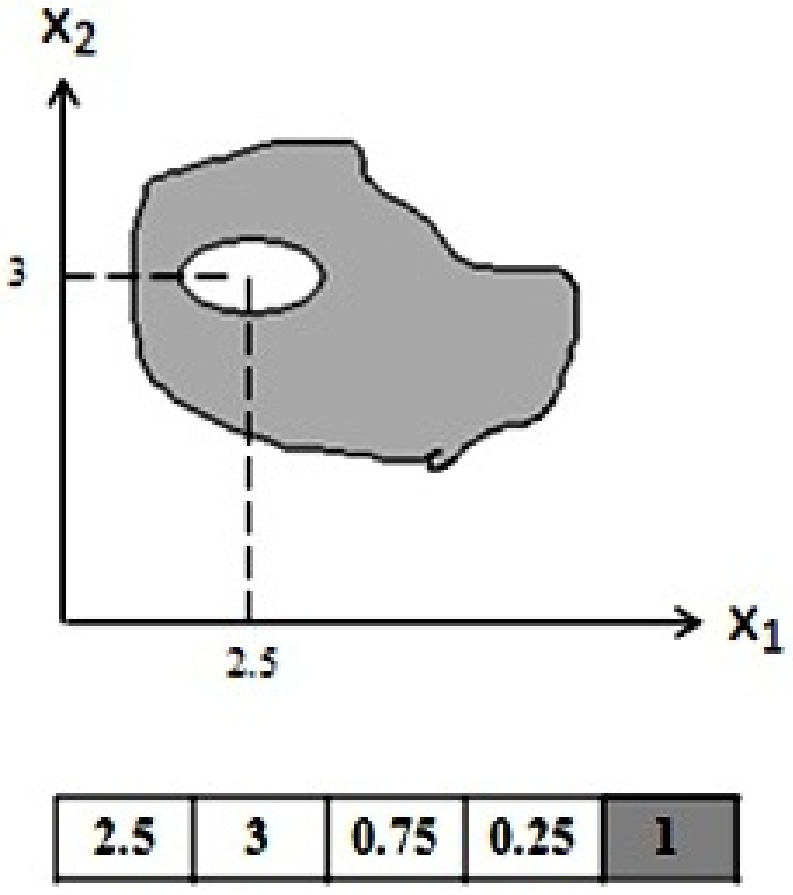}}
\begin{center}
\subfigure []{
\includegraphics[width=0.3\textwidth]{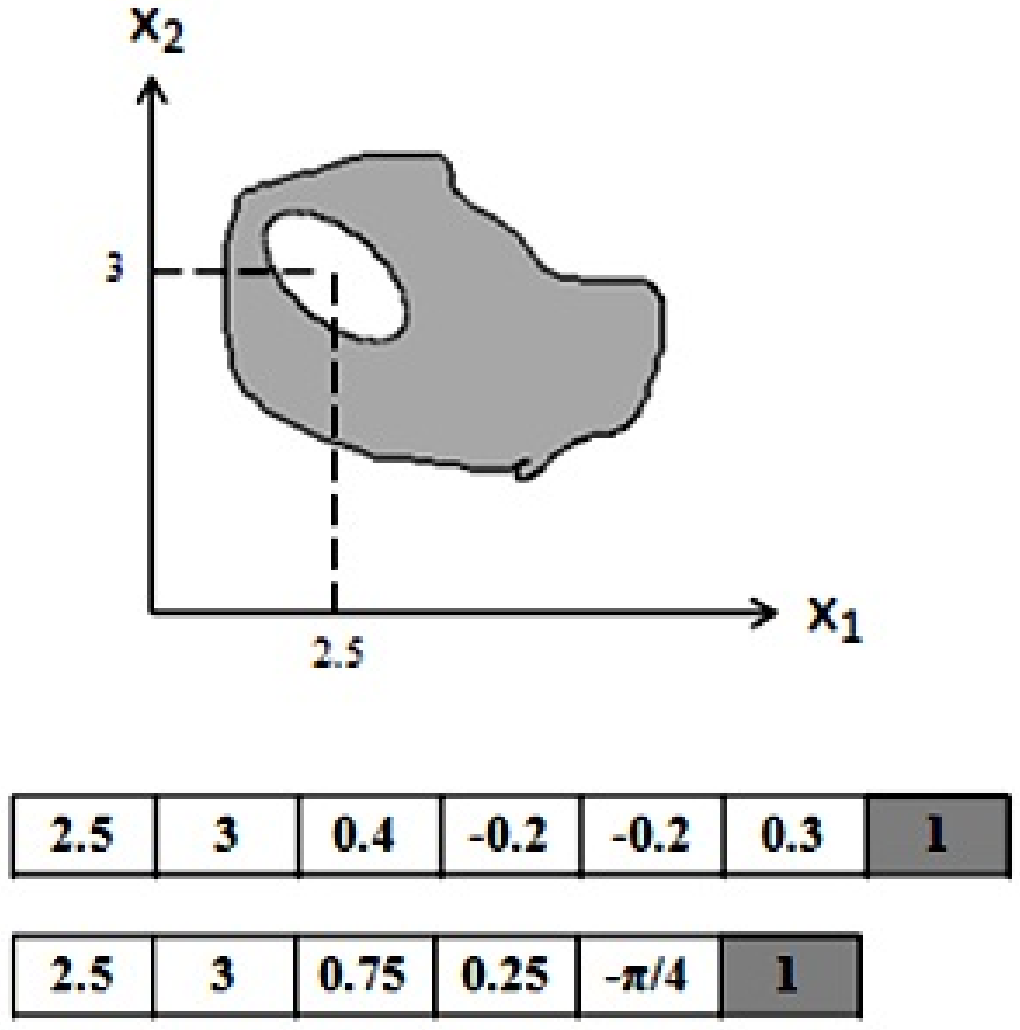}}
\end{center}
\caption{An example of the ellipsoidal based classifier in an arbitrary problem space. a) Visualization of the covered area by a hyper spherical classifier and its genotype. b) Visualization of the covered area by an axis-parallel hyper ellipsoidal classifier and its genotype. c) Visualization of the covered area by a general hyper ellipsoidal classifier and its genotypes in two cases; in the upper one, matrix $\Sigma$ is used to identify the rotation of the represented ellipsoid and in underneath case, the orientation angle is presented. The white cells identify the condition part and the gray cell identifies the action part of the corresponding classifier.}
\label{fig:3}       
\end{figure}

In all these structures, activation of each classifier in forming $[M]$ is determined by a Gaussian kernel function, applied to the distance between the current input and the center point. So, to find whether a classifier can match the current input $x$ or not, at first, the activation of the classifier \textit{$cl.ac$} is computed using Formula 1, 2 or 3 according to its structure. Then, the current input will be matched with a classifier if the activation of such classifier is greater than a threshold $\theta_m$. Formula 1, 2 or 3 are used when the condition of the classifier represents a hyper sphere $(C=\{\overrightarrow{m},\sigma)\}$, an axis-parallel hyper ellipsoid $(C=\{\overrightarrow{m},\overrightarrow{\sigma}\})$ , or a general hyper ellipsoid $(C=\{\overrightarrow{m},\Sigma\})$ respectively.

\begin{equation}
cl.ac=exp(- \frac{{\|x-m\|}^2} {(2\sigma^2 )}).
\end{equation}

\begin{equation}
cl.ac=exp(-\sum_(i=1)^n\frac{(x_i-m_i)^2}{(2\sigma^2 )}).
\end{equation}

\begin{equation}
cl.ac=exp(- \frac{(\overrightarrow{x}-\overrightarrow{m})^T \Sigma^T \Sigma(\overrightarrow{x}-\overrightarrow{m})}{2}).
\end{equation}

Promising results of these condition representation approaches reported in (Butz, 2005; Butz et al., 2006; Butz et al., 2008) showed an improvement in performance of XCS (or XCSF) in continues space problems. Especially, it is more profitable to use a more general condition structure while the dimensional dependencies of the problem are unknown. Although these representations can appropriately model oblique boundaries, but it must be noted that it might be not well-suited where decision boundaries of problem are axis-parallel for which case interval based representation can be applied effectively.

\subsection{Convex Hull Based Representation}
\label{sec:6}
Another approach that can be utilized to represent the condition of a classifier in the real-valued problems is using the convex hull concept as proposed in (Lanzi and Wilson, 2006). In this approach, condition part of each classifier comprises a set of points in the problem space that identifies a convex hull. In other words, each classifier depicts a convex region of the problem space and matches all problem instances which lie inside this region, as shown in Figure 4. As all the other geometric shapes can be approximated through a convex hull with sufficient number of points, the convex hull based representation is a general form of both previous representations. Besides, the convex hull representation has a fine ability to identify more complex regions due to its asymmetric shape.

\begin{figure}[!htb]
\begin{center}
\subfigure []{
\includegraphics[width=0.25\textwidth]{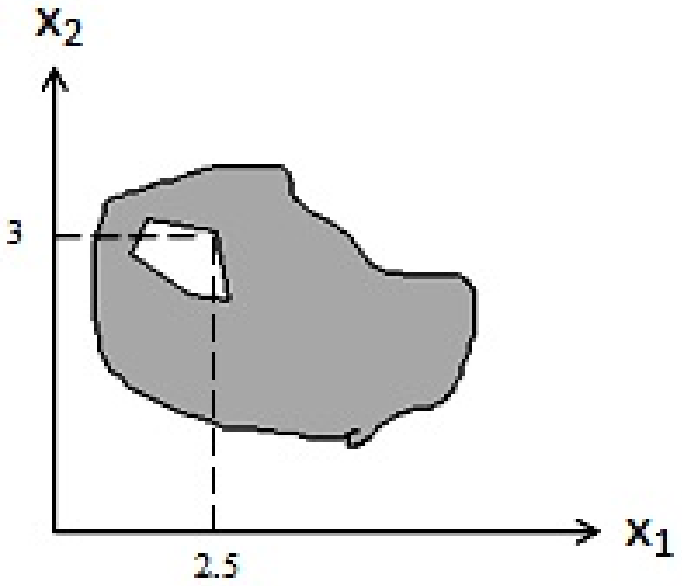}}
\subfigure []{
\includegraphics[width=0.4\textwidth]{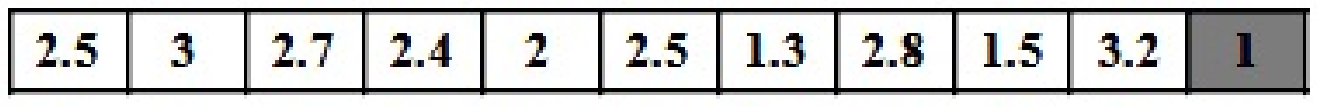}}
\end{center}
\caption{An example of a convex hull based classifier in an arbitrary problem space. a) Visualization of the covered area by a convex hull based classifier. b) The genotype of a classifier presented in (a) which defines the points to represent the desired convex hull. The white cells identify the condition part and the gray cell identifies the action part of the corresponding classifier.}
\label{fig:4}       
\end{figure}

Condition part of the classifiers can represent a convex hull in two manners; first, using a set of points. Second, each classifier can define a convex hull directly by presenting angles and radius of the convex hull in its condition part, instead of using a set of points. In both cases, number of convex hull vertices can be fixed or variable. In (Lanzi and Wilson, 2006), the authors discussed the influence of variable sized conditions and concluded that\textit{ '… represent arbitrarily complex convex regions but increase the complexity of the genetic search and also introduce the bloating phenomena that are typical with variable size representations …'}.

In (Lanzi and Wilson, 2006), the performance of XCSF based on convex hull representation was compared to the XCSF with interval-based representation. Results showed the fast convergence of system when the problem space is partitioned by convex regions. Also it was shown that XCSF with variable sized condition converges faster than a version of XCSF where condition part of classifiers consists of 10 or 15 points, but compared to 3 and 5 points, it converges slower.

\subsection{Fuzzy Logic Based Representation}
\label{sec:7}
In the rule based classifier systems, the comprehensibility and interpretability are two must considerable features (Hayes-Roth, 1985). Besides, fuzzy logic is one of the best known mechanisms providing such properties. There are number of approaches to use fuzzy logic and fuzzy set theory (Zadeh, 1965; Zadeh, 1973) as a technique for representing rules in Michigan-style LCSs, such as (Valenzuela-Rend$\acute{o}$n, 1991; Bonarini and Matteucci, 2007; Casillas et al., 2007; Orriols-Puig et al., 2008a). The main goal behind such efforts is combining the generalization capabilities of LCS with the fine interpretability of fuzzy rules to achieve an online learning system with more accurate, general and well understandable rule set. In the following, first, we briefly describe important works in this area. Second, a comprehensive description of notable approaches which try to embed fuzzy logic in knowledge representation component of LCS is provided. 

Early attempts to integrate fuzzy logic and learning classifier systems were proposed in (Valenzuela-Rend$\acute{o}$n, 1991; Nomura et al., 1998; Parodi and Bonelli, 1993). Valenzuela-Rend$\acute{o}$n in (Valenzuela-Rend$\acute{o}$n, 1991) introduced the first proposal of Learning Fuzzy-Classifier Systems (LFCSs), which is a Michigan style LCS and consists of fixed-size fuzzy-rule set and a fuzzy message list. Afterwards, several researchers have used the idea of fuzzy logic or fuzzy set theory into LCS to have an online fuzzy rule based system which can be used in many tasks (Nomura et al., 1998; Parodi and Bonelli, 1993; Furuhashi et al., 1994; Nakaoka et al., 1994; Velasco, 1998; Ishibuchi et al., 1999). Many of proposed systems were applied to the reinforcement learning and control tasks. As the initial LFCS framework was not completely coincident with reinforcement learning architecture, Nomura et al. (Nomura et al., 1998) improved the LFCS structure to be a true reinforcement learning technique. In (Parodi and Bonelli, 1993), Parodi and Bonelli defined an LFCS which can automatically learn the fuzzy relations, membership function and weights. Velasco (Velasco, 1998) designed a new extension of LFCS especially designed for fuzzy process controls and Ishibuchi et al. (Ishibuchi et al., 1999) proposed an LFCS for pattern classification. 

The research on this topic had been continued by the works of Bonarini (Bonarini et al., 2000; Bonarini, 1998; Bonarini, 2000; Bonarini and Matteucci, 2007). He addressed the classic "Competition versus Cooperation" problem in genetic fuzzy systems, (Bonarini, 1996; Bonarini and Trianni, 2001). He proposed a Michigan style LCS named ELF where the rule set is divided into subpopulations. To produce the correct action, the classifiers of these subpopulations cooperated whilst the classifiers in each subpopulation competed with each other. The behavior of ELF in overcoming some of issues of strength-based LCSs was verified by applying it to several reinforcement learning problems such as the coordination of autonomous agents. In (Bonarini, 2000; Bonarini et al., 2007), a general framework of learning classifier systems were introduced and later this framework was extended particularly for XCS called FIXCS in (Bonarini and Matteucci, 2007). In (Bonarini, 2000), the different components of this framework have been analyzed to be consistent with fuzzy models. In addition, some features are introduced for the sake of classifying LFCS proposals presented in the literature. 

Recently, in (Casillas et al., 2004; Casillas et al., 2007) Casillas et al. also used fuzzy logic in knowledge representation component of XCS to express rules in fuzzy format. There, the theoretical issues of applying fuzzy model in accuracy based learning classifier systems have been investigated. The proposed approach was named Fuzzy-XCS which is the first successful accuracy based fuzzy rule based system with generalization ability. Soon after, the Orriols-Puig proposal (Orriols-Puig et al., 2008a; Orriols-Puig et al., 2008b), named Fuzzy-UCS, extended this approach to be applied in supervised learning classifier system, i.e. UCS which is a derivation of XCS introduced in 2003 (Bernad$\acute{o}$-Mansilla and Garrell, 2003) for classification task in data mining. In following, a comprehensive description of well known fuzzy representation which is successfully used in LCS realms to produce accuracy based fuzzy rule based system such as Fuzzy-XCS and Fuzzy-UCS.

The main idea of using fuzzy logic in XCS as the knowledge representation tool is to represent the labels associated to fuzzy sets in the rule's structures and offer LCS a mechanism to evolve them. This mechanism must be consistent with fuzzy rules and it must also be able to learn a rule set in order to implement an input to output mapping where both input and output can be either real valued or nominal. In common manner, the rule set consists of fuzzy rules which are defined in disjunctive normal form (DNF) with the following structure:

\begin{equation}
\textit{\textbf{IF} $X_1$  is  $Ã_1$  and  ...  and  $X_n$  is  $Ã_n$  \textbf{THEN}  Y  is  B.}
\end{equation}

Where each input variable $X_i$ is described through a set of linguistic terms, $Ã_i  =\{A_{i1}  V ... V A_{il}\}$, represented by a disjunction (T-conorm operator) of $l$ linguistic terms. Meanwhile, each output variable is represented by a usual linguistic variable. For example, in (Casillas et al., 2007) to use this representation in classifiers of Fuzzy-XCS, a binary coding schema for the condition part and an integer coding schema for the action part were proposed. The number of bits considered in the condition part is equal to the total number of defined linguistic terms for all dimensions. For example, consider a problem with two dimensions where five linguistic terms are defined for both the first and second dimensions. So, the condition part of each classifier should have ten bits to codify these linguistic terms. The value of each bit indicates the appearance of the corresponding linguistic term. For each output variable, there is a gene in action part which denotes the index of utilized linguistic term. Below, there is an example to show how a rule can be codified by this representation:

\[\textbf{\textit {Terms for each input\slash output variable:}}\]
\normalsize{     vS [very small], S [small], M [medium],}\\
\normalsize{     L [large], vL [very large] }
\[\textbf{\textit{Fuzzy rule:}} \]
\begin{center}
\normalsize{ IF  $X_1$  is   \{M,L\}   and   $X_2$   is   S   THEN   $Y_1$   is   M  and   $Y_2$   is   L }
\end{center}
\[\textbf{\textit{Representation:}} \]
\normalsize{[ 0011001000  $\|$  23 ]}

Figure 5 shows what portion of input space can be covered by a fuzzy classifier. As it is shown, each dimension is partitioned according to its defining linguistic terms and the covering region is denoted by a dashed rectangle.

\begin{figure}[!htb]
\begin{center}
\subfigure []{
\includegraphics[width=0.45\textwidth]{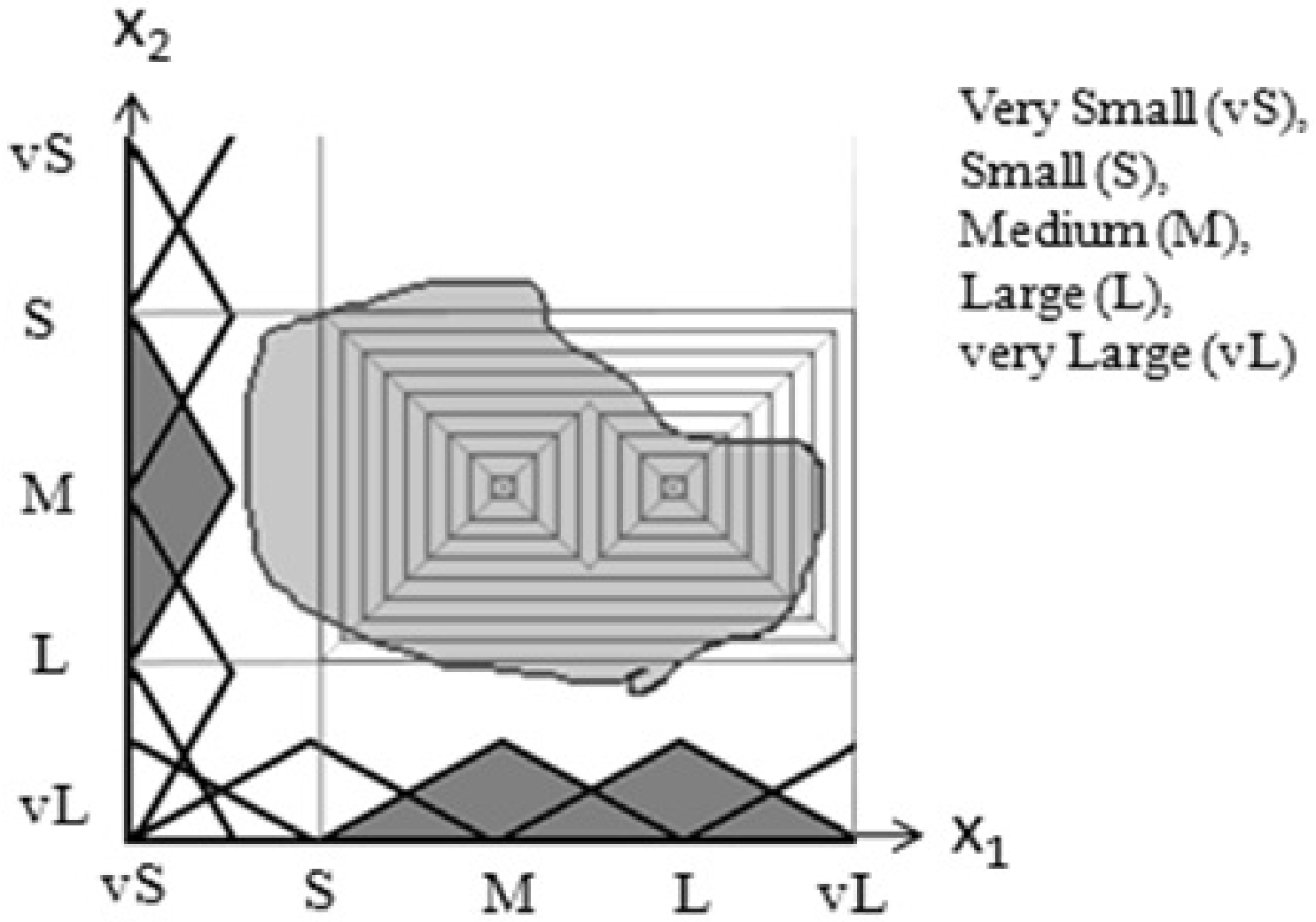}}
\subfigure []{
\includegraphics[width=0.4\textwidth]{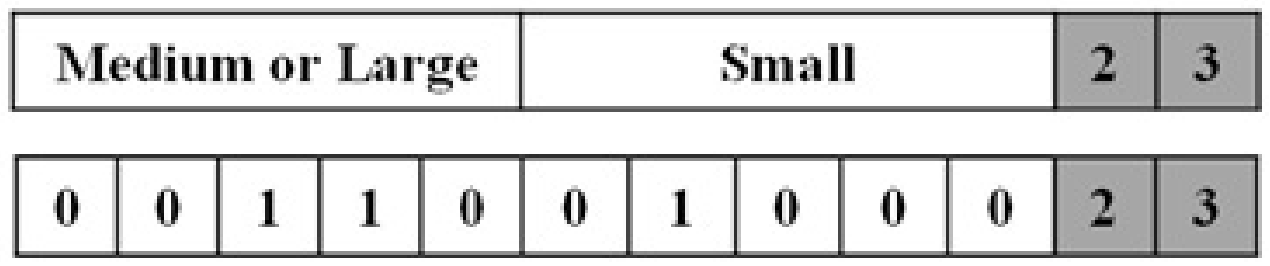}}
\end{center}
\caption {An example of a fuzzy classifier in an arbitrary problem space. a) Visualization of the covered area by fuzzy classifier. b) The phenotype and genotype of classifier presented in (a). The white cells identify the condition part and the gray cells identify the action part of the corresponding classifier}
\label{fig:5}       
\end{figure}

Rules with fuzzy structure entail several features making it useful as claimed in (Casillas et al., 2007): 
\begin{itemize}
    \item It offers an approach to handle mixed attribute.
\item It can handle missing values due to the natural capability of fuzzy logic in supporting the absence of some input variables.
\item This structure leads to have more compact description and the evolved rule set consists of fuzzy rules with different generalization degrees.
\item It is robust to noisy input.
\item It can produce the output classes with a certainty factor in classification problems.
\end{itemize}

Promising results are obtained by evaluating the system on some function approximation problems and a realistic robot simulation online learning and supervised learning problems. Results show that it can increase the compactness of evolved rule set which is accurate, general and co-adapted However, using fixed and predefined fuzzy sets cause a limited number of possible rules. So, they are not flexible enough to fit the data very well.  In order to overcome this issue, system must have an ability to modify the predefined fuzzy sets. 
In (Marin-Bl$\acute{a}$zquez et al., 2007; Marin-Bl$\acute{a}$zquez and Shen, 2008; Marin-Bl$\acute{a}$zquez and Martínez Pérez, 2008), linguistic hedges are employed to produce a new fuzzy set by modifying the original fuzzy set in interpretable manner where the original fuzzy set remains unaltered. By using linguistic hedges, the model has more freedom in domain knowledge representation and extraction. The research has demonstrated that better granularity of linguistic fuzzy modeling are achieved by using linguistic hedges as a technique to modify fuzzy sets and the inclusion of hedges causes improvement in the accuracy of the resulting system.

\subsection{First Order Logic Based Representation}
\label{sec:8}
Mellor proposal in (Mellor, 2005; Mellor, 2006) extended XCS to a new model named FOXCS where rules are represented based on the first-order logic. First-order logic is useful to improve the expressive power of XCS due to its ability to present the complex relationships among attributes of a task domain. The modified classifiers are in form of Horn clauses which consist of three parts; an action part, a condition part and a background part, that is \textit{action $\leftarrow$ condition,Background}. Condition part consists of a number of variables. Here, variables have a generalization role in XCS similar to the \# symbol in the ternary representation. However, variables can also be placed in action part. But, as a classifier should be defined in Horn clause form, its action part must consist of just one variable, i.e. atom. The background part can be empty. 

The main advantages of using the first order logic are; 1) it is facilitated to represent relational concepts which relate variables of action part to those of condition, like \textit{(A $\leftarrow$ ABB)} where $A$ and $B$ are variables of given problem. Obviously, the variable A which takes place in action part can also appear in condition part. To illustrate this property, there is an example for a \textit{Blocks World} tasks in Figure 6. 2) Another advantage is that rules can contain background knowledge which is a feature of many inductive logic programming (ILP) and rational reinforcement learning (RRL) systems, and can be helpful to solve these tasks effectively. 3) Mellor [Mellor, 2008] claimed that FOXCS as a RRL system is general, model free and "tabula rasa"  system. It is general due to no restriction in problem framework and "tabula rasa"\footnote[1]{Tabula rasa is the epistemological thesis that individuals are born without built-in mental content and that their knowledge comes from experience and perception.} because the initial policy can be left unspecified.
\begin{figure}[!htb]
\begin{center}
\subfigure []{
\includegraphics[width=0.3\textwidth]{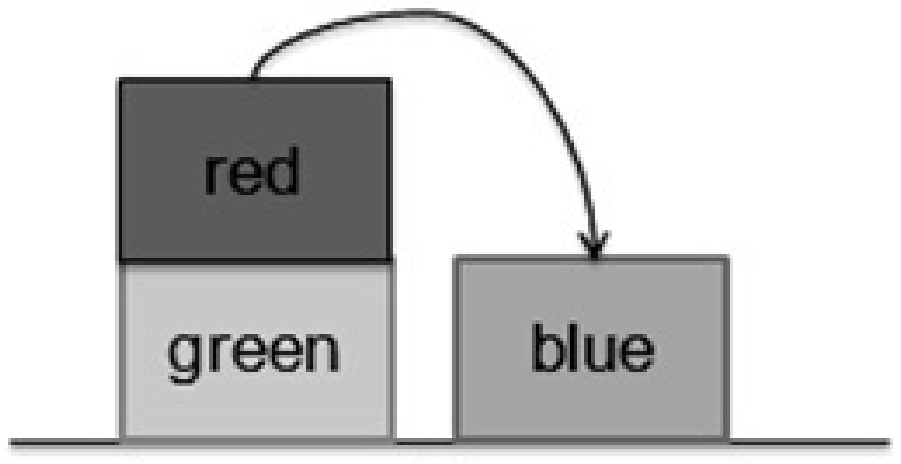}}
\subfigure []{
\includegraphics[width=0.45\textwidth]{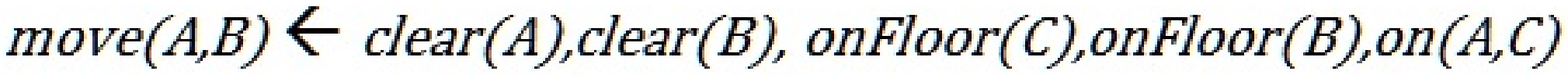}}
\end{center}
\caption {An example of a first order logic based classifier in a Blocks World problem. a) Current situation of problem environment. b) A classifier based on first order logic, which can match the current situation. $A$, $B$ and $C$ are problem variables and $clear(X)$, $onFloor(X)$, $on(X,Y)$, and $move(X,Y)$ are functions. This classifier will match current situation if the $A$, $B$ and $C$ variables are assigned to \textit{"red"}, \textit{"blue"}, and \textit{"green"} values.}
\label{fig:6}       
\end{figure}
The Results reported in (Mellor, 2005; Mellor, 2006; Mellor, 2008) show that the extended XCS can reach near optimal solution in the \textit{Poker} and \textit{Blocks World} problems which are not solvable by traditional XCS. Experimental results demonstrated that FOXCS can learn the optimal or near optimal policies where accuracy is comparable to the accuracy of many ILP algorithms in solving standard RRL tasks.

\subsection{Messy Code Representation}
\label{sec:9}
In (Lanzi, 1999), Lanzi introduced the messy version of XCS classifier system named XCSm in which the condition part of each classifier is defined by messy coding. The messy representation was firstly introduced by Goldberg in Messy Genetic Algorithm (Goldberg et al., 1989). In this representation, each gene is defined by a pair: \textit{(Gene Number, Allele Value)}. As it is not necessary to have an allele for each possible gene in a chromosome, messy chromosome can be underspecified. In XCSm, a variable sized set of messy genes composes the classifier condition in which Gene Number presents the tested sensor and Allele Value is a fixed length bit string that specifies the sensed input. For example, consider a maze environment with eight sensors. The food, obstacle and empty cells are codified as \textit{"11"},\textit{"10"}, and \textit{"00"} alternatively. Any input state with a goal or obstacle in its south position is matched by a messy gene like (S,"1\#"). Figure 7 shows an example of such classifier in Maze7 problem introduced in (Lanzi, 1998). As shown in Figure 7(a), the corresponding classifier can cover states marked by star in their center. With comparison to Ternary representation, the main advantageous of this representation is the independency of this representation's bits from the position of input sensors bits.
\begin{figure}[!htb]
\begin{center}
\subfigure []{
\includegraphics[width=0.2\textwidth]{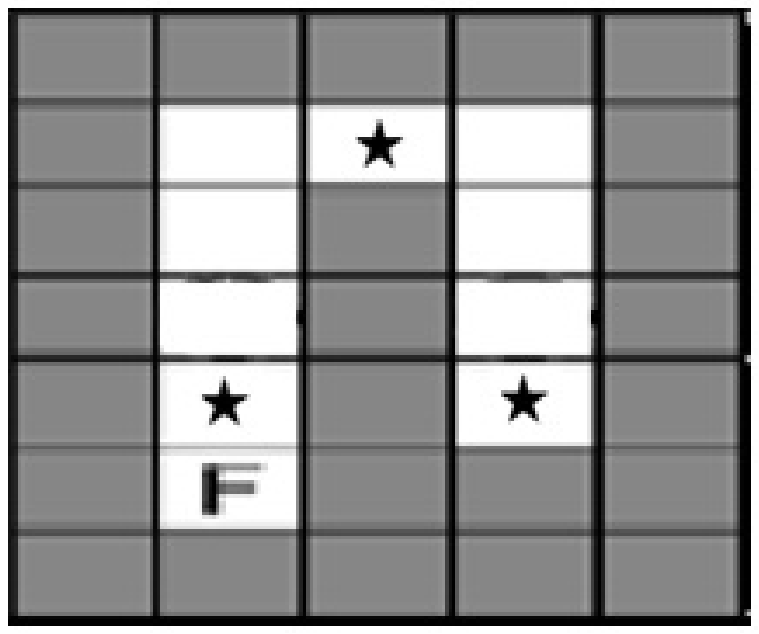}}
\subfigure []{
\includegraphics[width=0.2\textwidth]{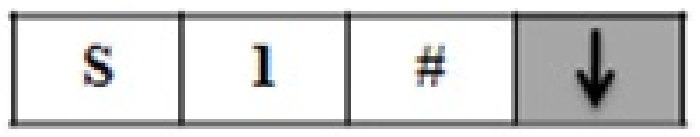}}
\end{center}
\caption {An example of a messy code based classifier in Maze7 (Lanzi, 1998) problem space. a) Visualization of the problem space. b) an arbitrary classifier. The matched states of environment states are marked in part (a) by a star in their center.}
\label{fig:7}       
\end{figure}
Experimental results in (Lanzi, 1999) showed that XCSm cannot reach the optimal solution. By analyzing the final population of classifiers, it is found that most of the evolved classifiers are over-general one. In order to cope with under-specification and over-specification of classifier condition, Lanzi developed XCSm by using high covering probability, extending mutation operator. He also suggested a new matching operator in which a condition matches current input if all its messy genes can match this input. As experiments notified, this new enhanced XCSm can learn an optimal solution.
\subsection{GP-like Representation}
\label{sec:9}
The representation methods which represent a tree form as GP (Koza, 1992), an expression as BNF\footnote[2]{Backus Normal Form or Backus-Naur Form}  grammar (Chomsky, 1956), and the other similar structures are known as GP-like representation. Firstly, using this representation in XCS was exploited by Lanzi in (Lanzi and Perrucci, 1999; Lanzi, 2001a). There, a system named XCSL was proposed that the condition parts of its classifiers have a general purpose representation, namely lisp s-expressions. The conditions were defined by the BNF grammar as shown in Figure 8.
\begin{figure}[!htb]
\begin{center}
\includegraphics[width=0.4\textwidth]{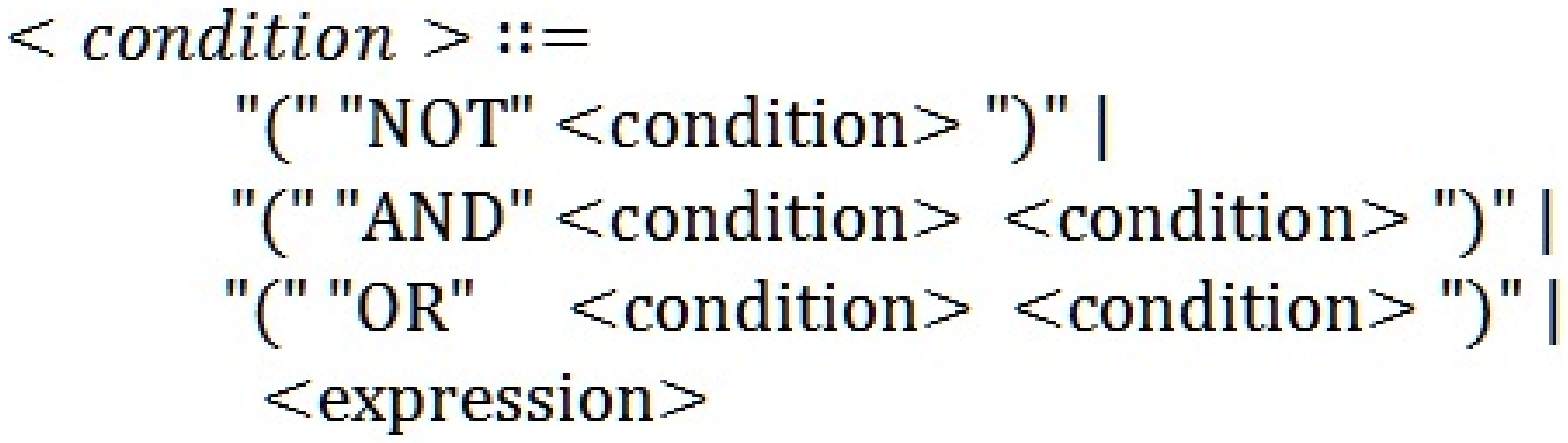}
\end{center}
\caption {The BNF grammar generating the overall structure of classifier conditions.}
\label{fig:7}       
\end{figure}
The non-terminal symbol \textless expression\textgreater  in BNF form must be defined in order to use this lisp like representation for a given problem.

To evaluate a classifier in the matching phase, the input data of XCSL must be represented as a string of attribute-value pairs and the terminal symbols presented in the classifier must be replaced with their actual values of corresponding attributes. Since the classifiers in XCSL have GP like structure, the genetic operators work as they do in traditional Genetic Programming (Koza, 1992). 

As shown in (Lanzi, 2001b), XCSL tends to produce many overgeneral classifiers in population, so the learning performance of XCSL would be influenced in some cases. Besides, due to the bloat phenomenon (Langton et al., 1996; Soule et al., 1996), which is common in any GP like system, it is almost impossible to analyze what kind of classification model is developed by XCSL. To overcome the former problem, a final condensation phase was added into the system to extract a minimal subset of classifiers as the final solution. It extracts a compact solution from many overlapping classifiers instead of extracting all best classifiers from the population. The promising results suggested that XCS with symbolic based representation might be an interesting approach to extract useful knowledge from data.
In (Lanzi, 2003), an extension to XCS was proposed by adding a stack-based representation into the classifiers' condition which was a linear program expression in Reverse Polish Notation (RPN) form. So, each condition consists of a sequence of tokens; each token can be a constant, a function or a variable which can be assigned to the corresponding input attribute value. There, a stack-based genetic programming, introduced in (Perkis, 1994), is used to mutate and recombine the classifiers. The reported results are quite interesting. Since genetic operators do not take into account any information about the structure of the operators defined in RPN and their arity, syntactically incorrect conditions can be easily generated. Consequently, the search space of the feasible solution would be even more highly redundant in comparison with XCSL.

Wilson has explored the use of gene expression programming (GEP) within XCS (Wilson, 2008) named GEP-XCS. The main aim of using GEP translation in XCS was not only to fit the environmental regularities better than rectilinear but also to produce a well understood rule set. The condition part of each classifier in GEP-XCS is expressed with an expression tree which can be evaluated by initializing the tree's terminals with the corresponding input attributes values. So, in the process of forming $[M]$, the obtained value of applying the presented expression in each classifier on the current input is compared with a predetermined threshold. If the obtained value exceeds this threshold, corresponding classifier will be added to the $[M]$. 

The most important feature of using GEP translation is that every chromosome generated in GA cycle is syntactically valid. In addition, the size of expression tree in GEP representation is limited by the fixed chromosome size therefore the bloat phenomenon which usually happens in GP does not occur. GEP has further important properties such as combining several genes into a single chromosome and the concurrent evolution ability which are disregarded in GEP-XCS. Figure 9 illustrates how a GP-like classifier can partition the problem space. In this example, the classifier presents an expression tree form shown in Figure 9 (b). According to the different types, namely using GP or GEP, to encode the phenotype of the classifier, the expression tree can present in two different ways shown in Figure 9 (b). As mentioned, the condition part of classifier in GEP based representation must have fixed size; here it is assumed that the length of condition part of each classifier must be fifteen. Hence, since the corresponding tree form only needs seven genes to be encoded, in GEP based representation the additional genes are initialized with terminal variables.  But, in the GP representation the classifier can have variable length so the length of classifier condition part can be equal to the number of nodes of the expression tree.
\begin{figure}[!htb]
\begin{center}
\subfigure []{
\includegraphics[width=0.3\textwidth]{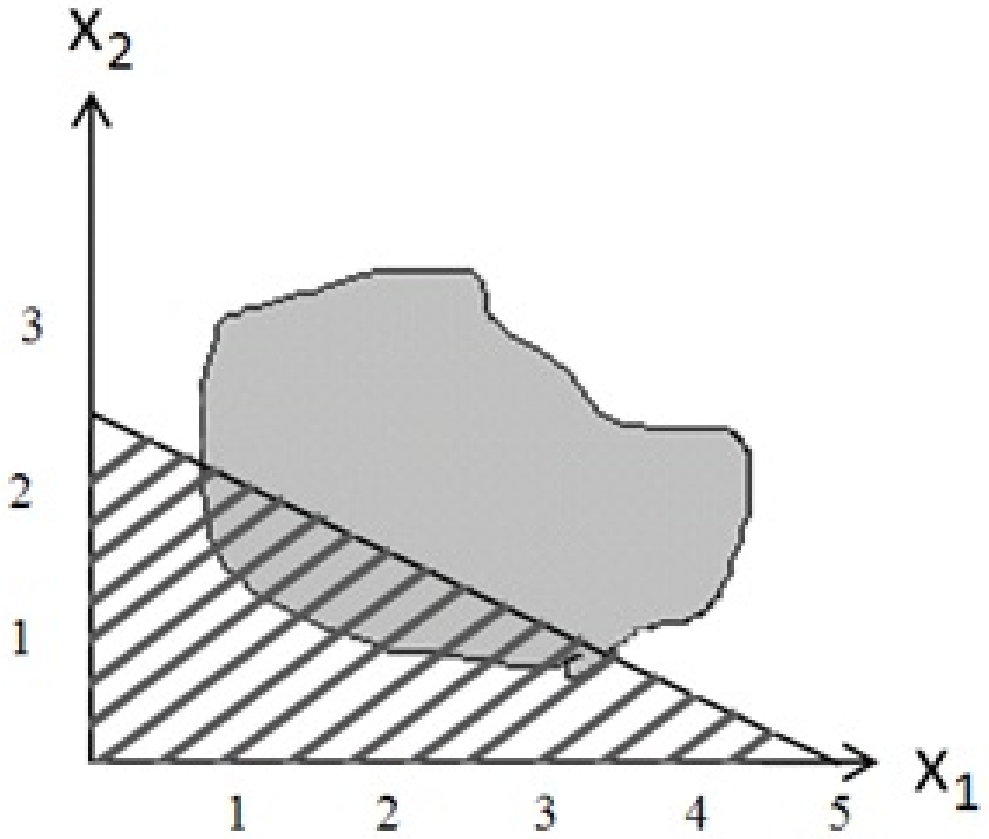}}
\subfigure []{
\includegraphics[width=0.45\textwidth]{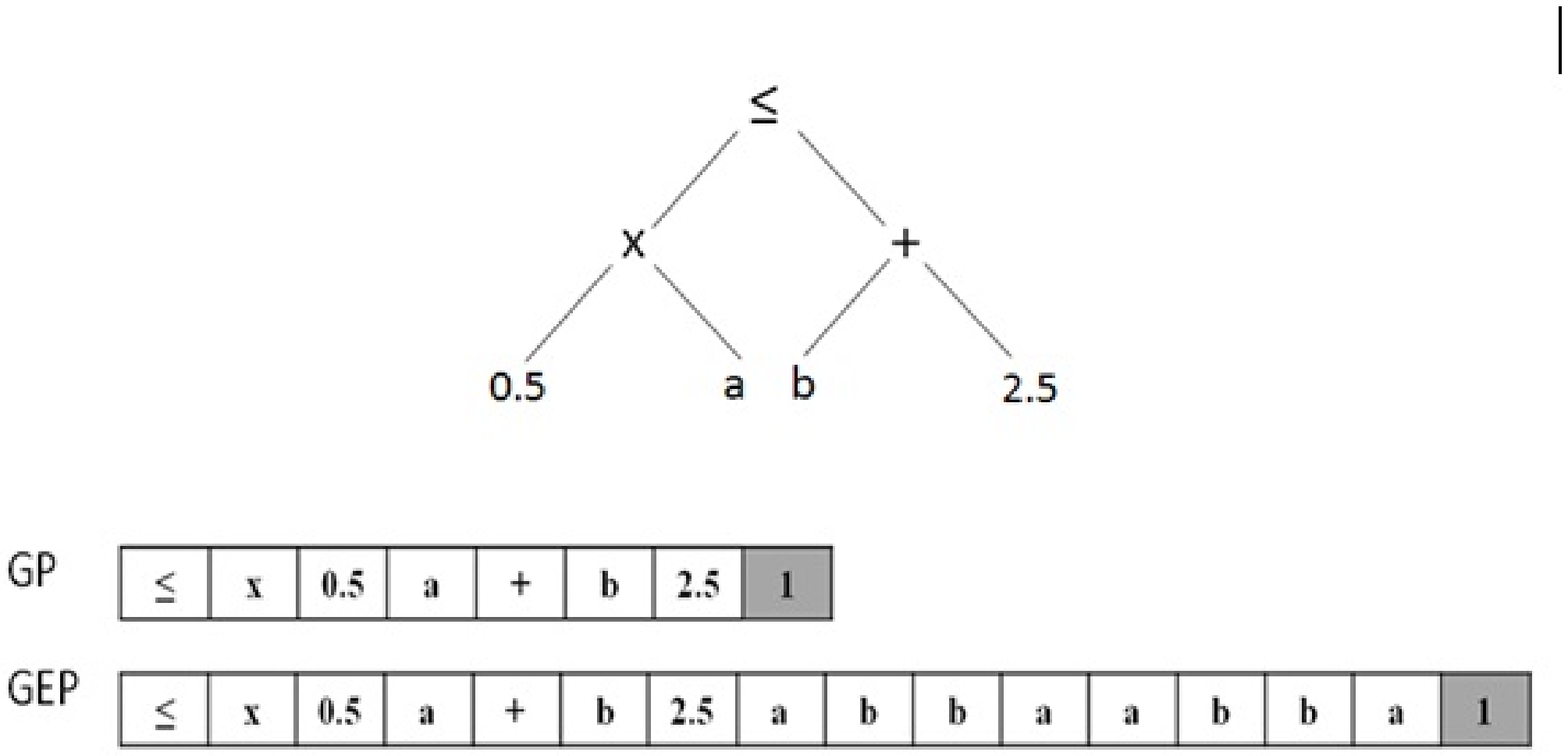}}
\end{center}
\caption {An example of a GP-like classifier in an arbitrary problem space. a) Visualization of the covered area by a GP-like classifier. b) The phenotype and genotype of classifier presented in (a).}
\label{fig:9}       
\end{figure}
Adding GEP to XCS as a condition representation has three main advantages; (1) each classifier has functional condition, with fine ability to fit environmental regularities, (2) the applied genetic operators can be quite simple because they are dealing with linear chromosomes instead of expression trees. Also, there is no need to verify the validation of produced offspring because genetic operators applying on GEP chromosomes always produce valid chromosomes and (3) the most attractive property which makes GEP based representation more desirable than other GP-like representations, is fixed size of expression tree. The reported results showed that using such representation can achieve good performance and gives greater insight into the environment's regularities but it leads fairly slower evolution and the evolved rule set was not compact.

Recently, an investigation (Preen and Bull, 2009) was done in using a discrete dynamical system representation within XCS which is termed Dynamical Genetic Programming (DGP). DGP-XCS used a graph based representation for the condition of each classifier, wherein each node is a Boolean function. In other words, each classifier condition presents a Random Boolean Network (RBN) with $N$ nodes which is equal to sum of the inputs $I$ and the outputs $O$ plus one $(N=I+O+1)$. The first connection of each input node is assigned to the corresponding locus of such input and other connections are set at random. The condition part of each classifier presents a RBN with its true table and the connections of each node. This new system is tested on two most common benchmarks, the multiplexer and the maze navigation problem. According to the obtained results, it can be concluded that it is possible to design an ensemble of RBN by using XCS which is able to solve a computational task under RL scheme.

\subsection{Neural Networks Based Representation}
\label{sec:9}
Larry Bull and Toby O'Hara proposed an accuracy based neural classifier named X-NCS and a Neuro-fuzzy classifier system called X-NFCS in (Bull and O'Hara, 2002). In X-NCS, both condition and action parts of each classifier present a single and full connected neural network like multilayer perceptron (MLP) (Bishop, 1995). In the condition part of each classifier, the weights of such small neural networks are concatenated together and can be evolved under GA mechanism in XCS. All the neural networks presented in all classifiers have the same number of nodes in their hidden layers. Besides the number of output nodes is equal to sum of the number of possible actions and one extra output node named \textit{'not match-set member'}. Since all rules can see the input state, this extra output node is considered to signify whether the corresponding classifier is a member of the $[M]$ or not. Each classifier whose \textit{'not match-set member'} node does not have the highest output value can be a member of the $[M]$. The winner action of each classifier is the one whose corresponding output node has the highest activation.

X-NCS had been tested on two tasks; a 6-bit multiplexer which is a single step task and to solve a maze named \textit{Woods2} which is a multi step problem. The obtained results showed that X-NCS could solve both tasks. In addition, X-NCS components and parameters were modified to work as a function approximator. Its performance had been examined with root-mean-square. As expected, the results showed that the most accurate solution is achieved when one classifier can cover the whole input space. To extend X-NCS to Neuro-fuzzy version (X-NFCS), radial basis function (RBF) is used due to its similarity to the fuzzy rule based system (Jang and Sun, 1995). The discovery component in X-NFCS evolves RBFs as it does in X-NCS for MLPs.

One of the main advantages of this scheme, namely using neural network based representation, is its ability to be applied on problems with continuous action space. The promising results showed that both X-NCS and X-NFCS can be used in more complex tasks including both continuous and discrete action space and also it is applicable to both single step and multi step problems. To address the slow convergence in X-NCS, in (O'Hara and Bull, 2005) a hybrid technique was proposed to speed up the learning using Memetic Algorithm (MA) in XCS which augments the GA search with a local search (Moscato, 1989).

In (Howard et al., 2008), the principle of constructivism and self-adaptation was explored within XCS and XCSF. There, those systems were applied on well-known maze problems and the reported results showed using self adaptive operators, neural constructivism, and prediction computation could be an improvement in the realm of XCS. These systems can optimally perform in complex and noisy environments. Overall, the experiments of these mentioned papers and the similar ones showed that X-NCS is able to model the environment in simple robotics applications (Hurst and Bull, 2004) and also can solve several maze problems (Bull and O'Hara, 2002; O'Hara and Bull, 2005; Howard et al., 2008).
It must be mentioned that the complete replacement of a rule by a neural network makes X-NCS to lose its main advantage as a rule based systems, that is, the potential ability to produce a well understood rule set. To overcome this issue, a new system named NLCS was proposed in (Dam et al., 2008). It is an extension of UCS which is driven for classification task in data mining problems. In (Dam et al., 2008), neural network is just used to identify the proper action of classifiers. The aim of such usage is to produce smaller evolved rule set while it still maintains or even improves the predictive accuracy. So, each classifier consists of two parts: the condition part which can be encoded by any existing representation method such as interval predicates (which is also used in (Dam et al., 2008)) like messy code and GP-like, and the action part which contains a neural network. For better understanding, Figure 10 shows the structure of a classifier in a two-dimension problem and how such classifiers will partition the problem space.
\begin{figure}[!htb]
\begin{center}
\subfigure []{
\includegraphics[width=0.45\textwidth]{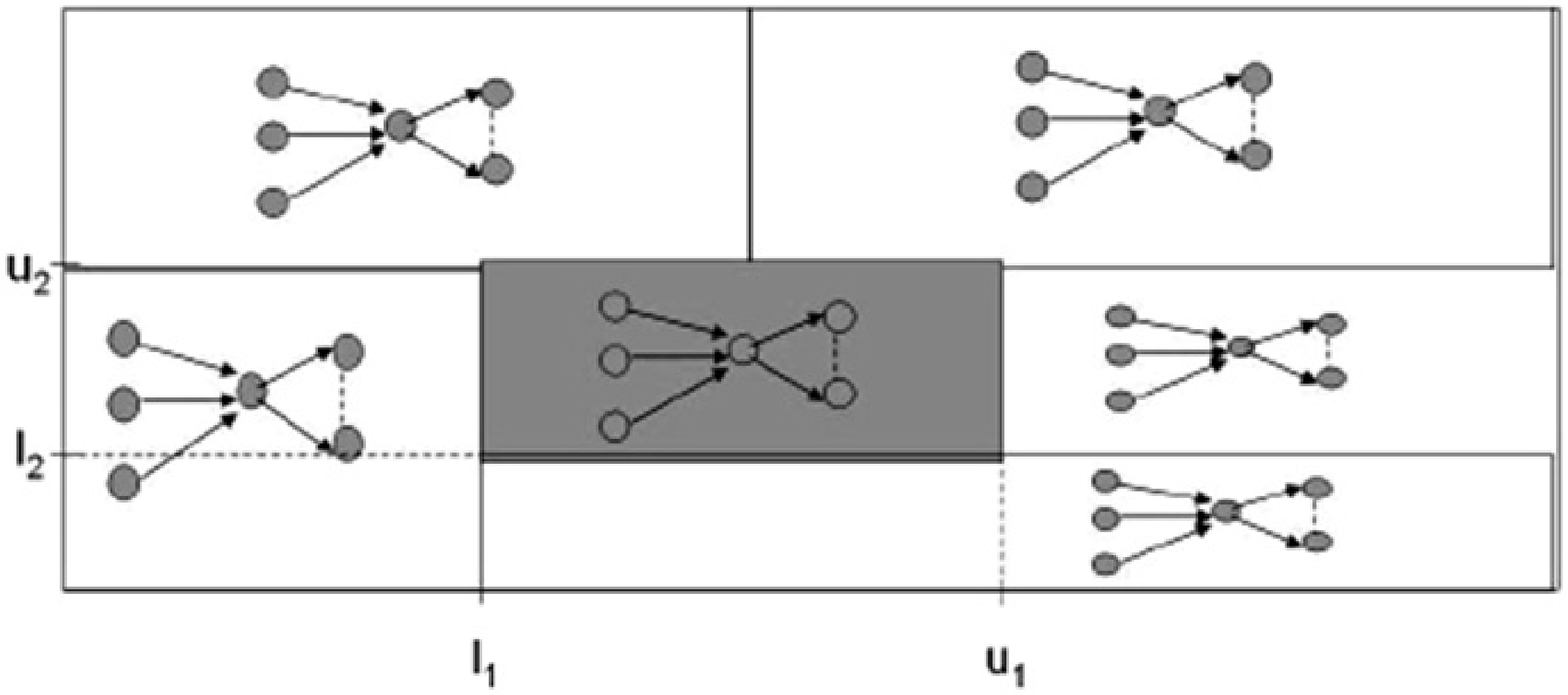}}
\subfigure []{
\includegraphics[width=0.4\textwidth]{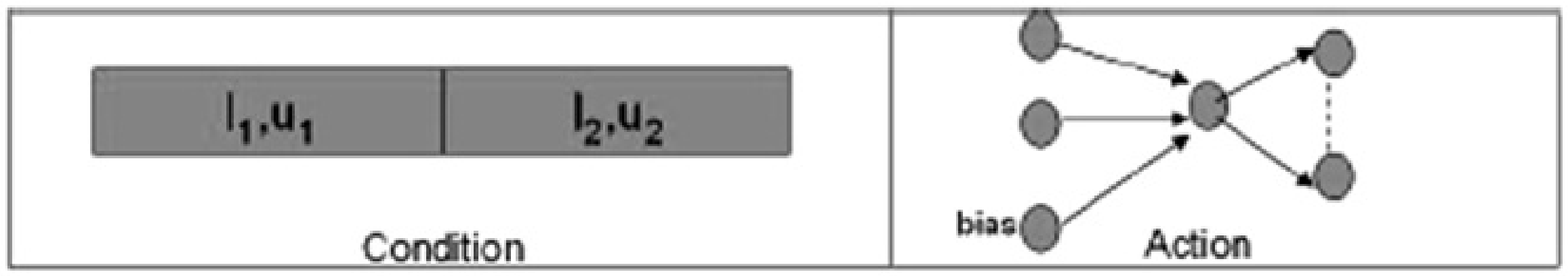}}
\end{center}
\caption {An example of a classifier of NLCS in an arbitrary problem space directly drawn from (Dam et al., 2008). a) Visualization of the covered area by NLCS's classifiers. b) The genotype of the classifier highlighted in (a) which has a neural network in its action part.}
\label{fig:10}       
\end{figure}
As results showed (Dam et al., 2008), defining neural network in action part allows XCS to produce more general classifiers because their condition part can cover large area of the problem space with more general hyper rectangle. So, NLCS can achieve better performance in classification problems with less number of evolved rules rather than UCS.

\subsection{Tile Coding Based Representation}
\label{sec:10}
Lanzi et al. had extended XCSF with tile coding prediction (Lanzi et al. 2006). Tile coding is one of the most common and successful methods to tackle complex environment in reinforcement learning realm (Sutton and Barto, 1998). In tile coding, the problem space is mapped into a set of overlapping tilings; each tiling partitions the input space into a set of nonoverlapping hyper-rectangles named tiles. Classifiers in XCSF with tile coding prediction have two additional parameters; the number of tilings and their resolution. The extended system can adapt these parameters through GA. Also the usual linear prediction function in XCSF is replaced with a tile coding approximator. So, the weight vector of each classifier contains the parameters related to each tile. Figure 11 shows how a classifier with two tilings partitions the input space and which tiles (states) could match the specified input.
\begin{figure}[!htb]
\begin{center}
\includegraphics[width=0.3\textwidth]{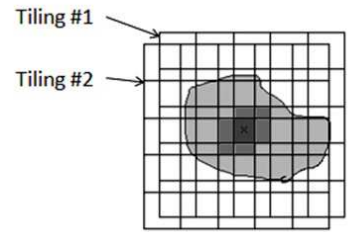}
\end{center}
\caption {Partitioning the problem space with a tile coding based classifier which suggests two way of tiling.}
\label{fig:11}       
\end{figure}
The new extended XCSF was tested on three multistep problems taken from the reinforcement learning literature: the 2D Gridworld (Boyan and Moore, 1995), the puddle world (Boyan and Moore, 1995), and the mountain car (Sutton, 1996). Indeed, such XCSF evolves an ensemble of tile coding approximators, each one on the problem subspace, instead of the typical monolithic approximator. The reported results showed that XCSF with tile coding can always reach an optimal solution and converge faster than XCSF with linear approximation.

\section{Relative Strength and Weakness of Knowledge Representation Techniques }
\label{sec:11}
Each of knowledge representation techniques discussed in previous section its unique strengths but also weaknesses. For solving a given problem effectively by a rule based system such as LCS and XCS in particular, it is important to know which knowledge representation technique is the well suited for identifying the problem regularities. In other words, the first step in solving a problem is identifying the problem area to choose a proper representation which can cover whole problem space completely and embrace its complexity effectively. In this section, we present a few simple problem settings and analyze the relative strength and weakness of each discussed technique in handling such problems. Let us consider the following categories of problems. It is worth mentioning that we name each category based on the decision boundaries shapes and the complexity of the problem regularities. 

\textbf{Theoretical Problems:} the main aim of this kind of problems is analyzing the behavior of the learning system in solving a problem without facing the complexity of environments. In other words, in these problems researchers try to choose a simple environment with specific properties. Hence, simple techniques for knowledge representation such as ternary for bit string inputs, and, interval based representation for real inputs could be a good choice.

\textbf{Orthogonal Problems:} in these problems the decision boundaries or regularities which are interesting to be modeled are parallel with the main axises. Figure 12 shows checkerboard problem, a sample problem of this problem category. 

\begin{figure}[!htb]
\begin{center}
\includegraphics[width=0.2\textwidth]{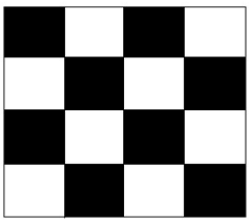}
\end{center}
\caption {Checkerboard problem: a classification problem data set. The positive instances belong to white region and the negative instances come from black region. This data set has axis-parallel boundaries.}
\label{fig:12}       
\end{figure}
Most of discussed knowledge representation techniques will handle this kind of problems. But interval based representation is well suited because it is a simple and popular technique and also it might be easier to apply compact rule set methods on the evolved rule set. In other words, since classifiers with interval based representation partition the problem space by defining axis-parallel hyper-rectangles of different sizes, this representation has an implicit ability to cover axis-parallel boundaries. But it is important to note that those problems with not only axis-parallel regularities but also continuous action can be handled using other representations such as neuro-fuzzy and neural network based representation. On the other way round, interval based representation will be able to handle such problems with continuous action if another independent method is used to produce a continuous action.

\textbf{Oblique Problems:} this category contains problems with oblique decision boundaries. Figure 13 shows several specimens of the recent type.

\begin{figure}[!htb]
\begin{center}
\subfigure []{
\includegraphics[width=0.19\textwidth]{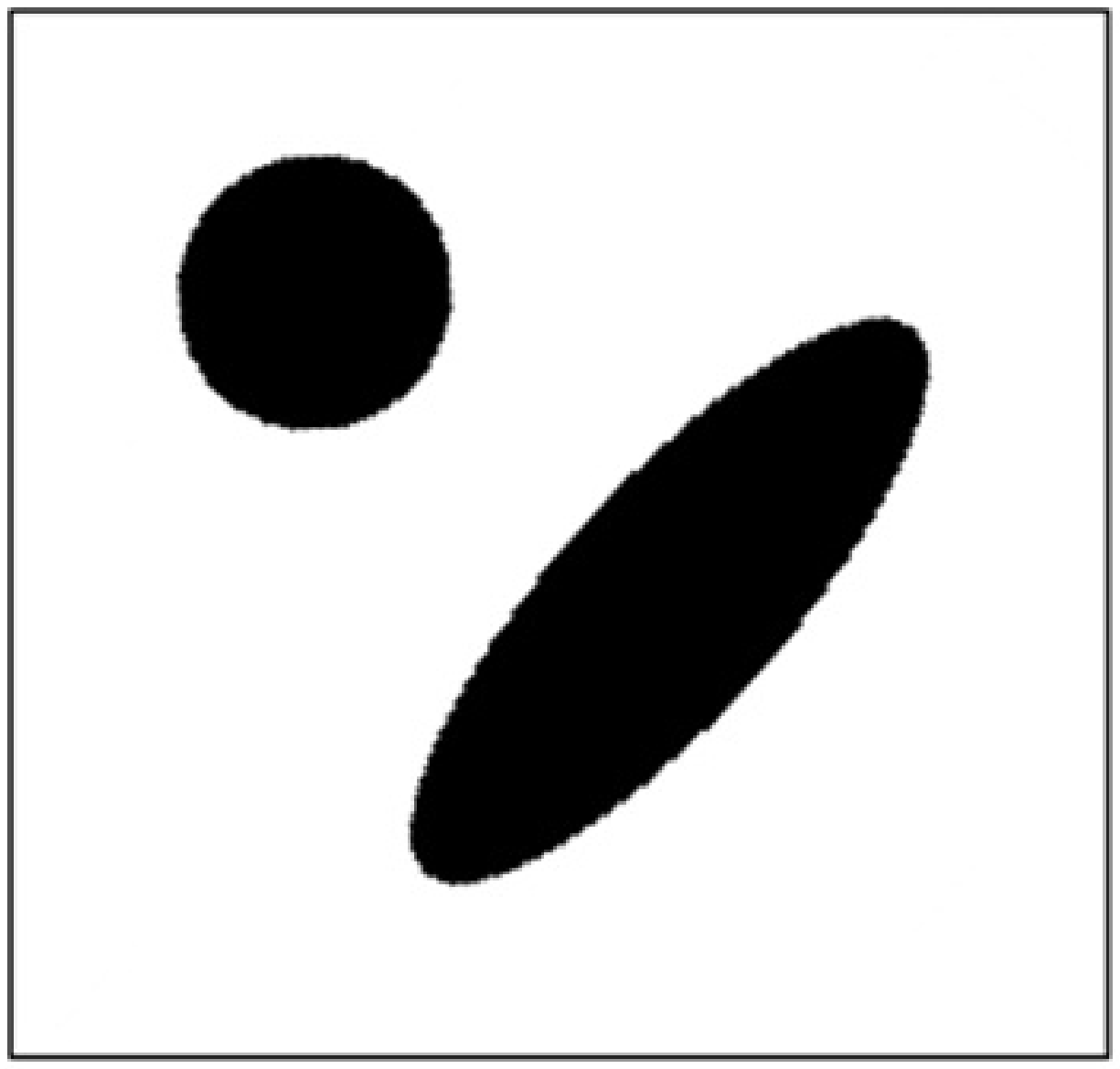}}
\subfigure []{
\includegraphics[width=0.18\textwidth]{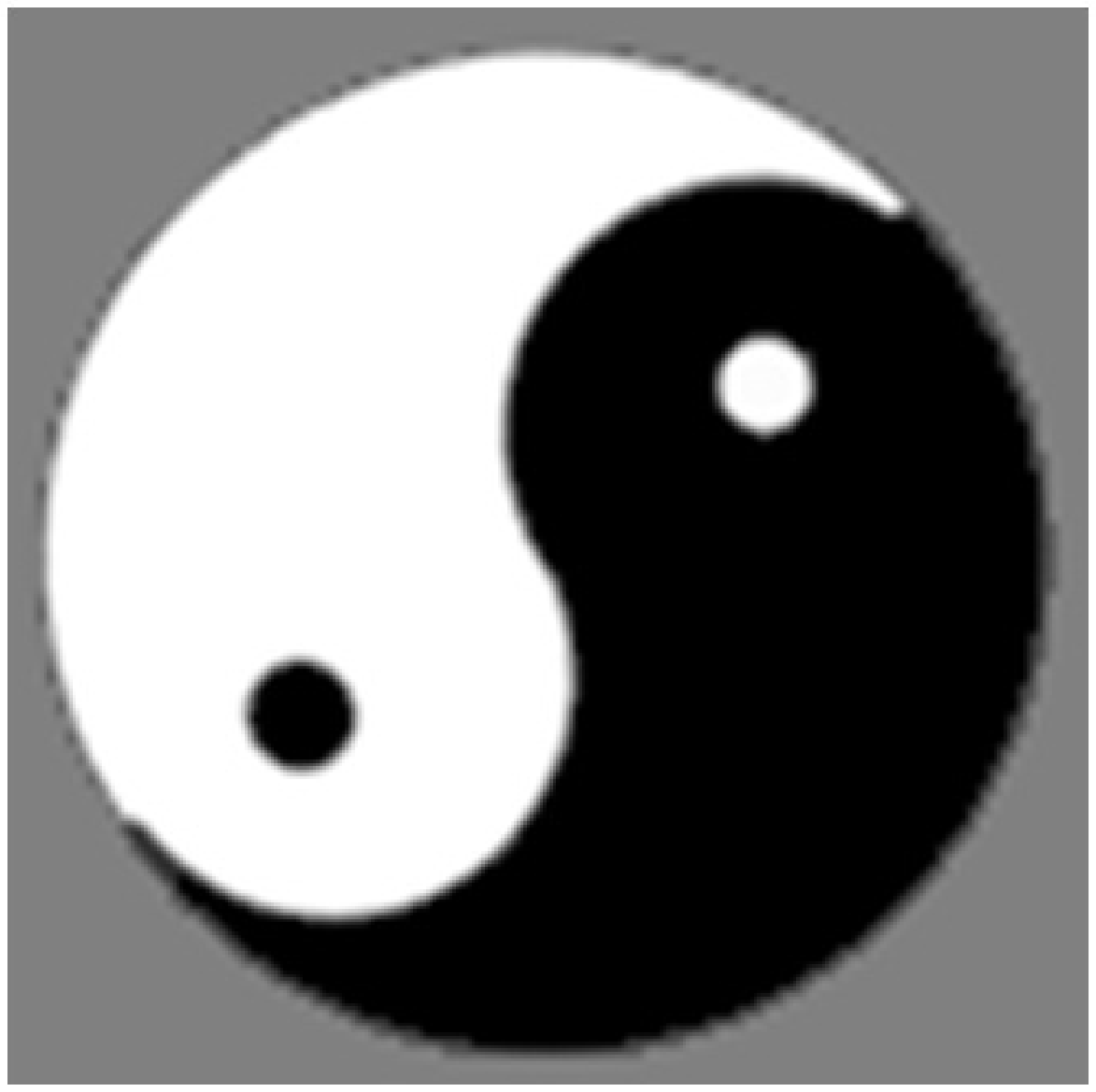}}
\subfigure []{
\includegraphics[width=0.18\textwidth]{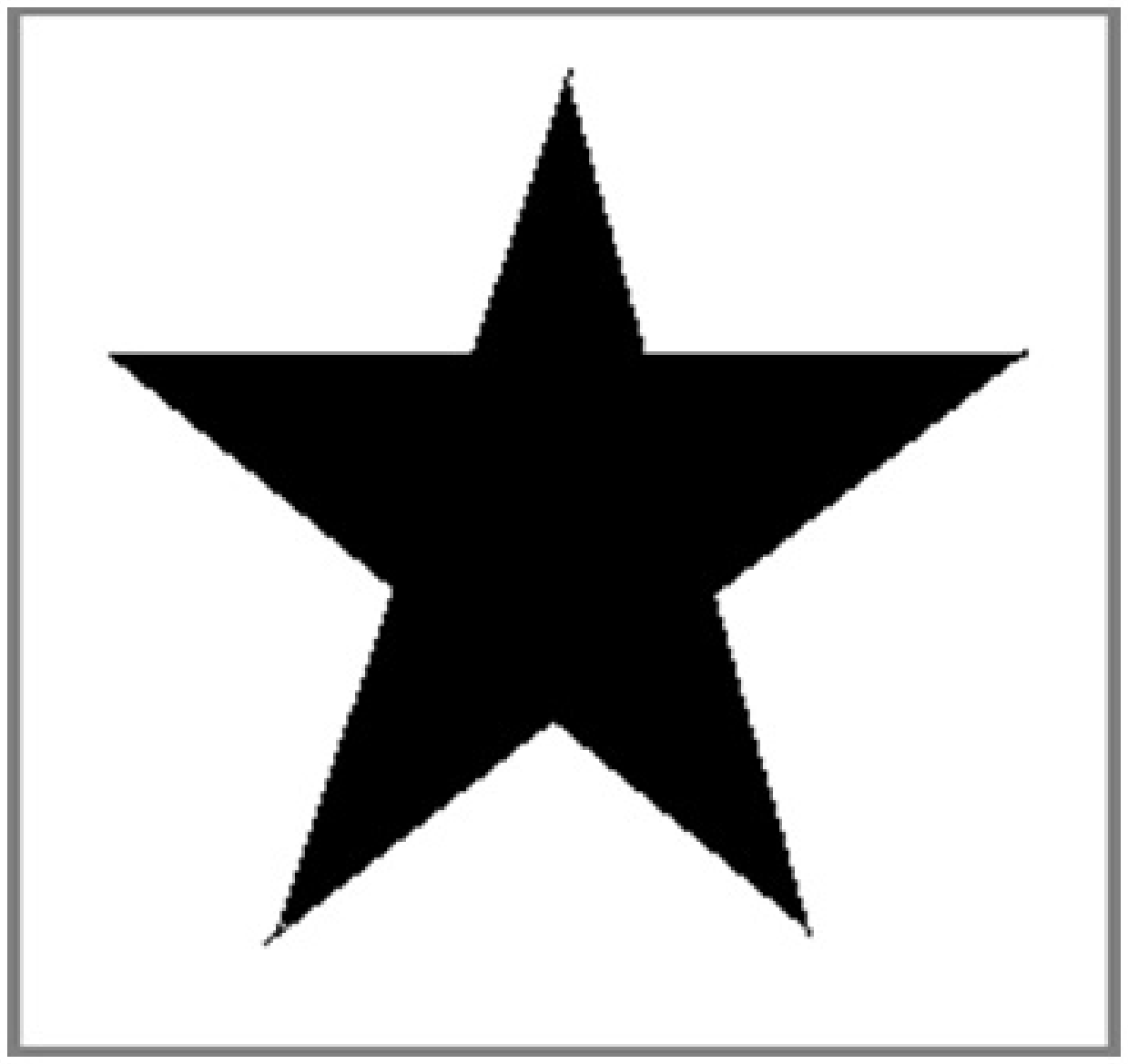}}
\end{center}
\caption {Sample problems of Oblique Problems. Black region belongs to negative class and white region belongs to positive region. The name of these problems are; (a) Cycloid, (b) Tao (Llor$\acute{a}$ and Guiu, 2001) and (c) Pentagram (Ji and Dasgupta, 2004).}
\label{fig:13}       
\end{figure}

If we have enough information about the regularities of problems, it is better to use techniques which can intuitively fit the boundaries properly. To elucidate, consider Cycloid and Tao data sets which have oblique cycloidal boundaries. A proper choice for such boundaries might be ellipsoidal based representation due to its natural cycloidal shapes. On the contrary, convex hull based representation can model the cornered decision boundaries of Pentagram data set adequately due to providing asymmetric shape and having higher flexibility in covering complex areas.

\textbf{Heterogeneous Problems:} This category contains the problems in which have complex regularities or in other world have heterogeneous decision boundaries. To handle such problems successfully as if the decision boundaries can be modeled more precisely, obviously one has to choose a technique among more complex representation with ability to cover complex problem spaces. Since knowledge representation techniques such as GP-like and neural network based representation are general purpose and able to represent arbitrary regularities; they are also the methods of choice when the problem has not only complex regularities but also unknown dimensional independencies.

\begin{table*}[htbp]
\caption{Existing approaches for knowledge representation in XCS with their main advantageous, disadvantageous, and the corresponding problem domains.}\label{conclusion}
{ \scriptsize
\setlength{\tabcolsep}{10pt}
\setlength{\extrarowheight}{2.5pt}
\begin{tabular}{|p{3.5cm}|p{4cm}|p{4cm}|p{4cm}|}
\hline
 \textbf{Rule representation \newline
(Systems)} 
&   
\textbf{Advantageous} 
\newline
&  
\textbf{Disadvantageous} 
\newline
&  
\textbf{Problem Domain}  
\\ \hline
\textbf{Interval Based Representation}
  
 (XCSR, XCSI)
 &\begin{itemize}
    \item Applicable to real valued   environments.
    \item Simple and easy to implement.
    \item Easy to analyze.
    \item Well-suited in problems which contain axis-parallel boundaries.
\end{itemize} & \begin{itemize}
    \item Difficult to match oblique decision boundaries.
    \item Limited to particular and symmetric shape.
\end{itemize} & \begin{itemize}
    \item Real-Valued/ Integer-Valued problems (Wilson, 2000a)
    \item Data Mining (Wilson, 2001b; Wilson, 2001; Bernad$\acute{o}$-Mansilla et al., 2001; Gao et al., 2005; Gao et al., 2007; Gao et al., 2006)
    \item Function Approximation (Wilson, 2002; Hamzeh and Rahmani, 2005; Hamzeh and Rahmani, 2007)
    \item Robotics (Hurst and Bull, 2004)
    \item Medical Data Mining-Ensemble Learning (Gao et al., 2007)
    \item Classification - Large Data Sets (Llor$\acute{a}$ et al., 2007,Dam et al., 2005b)
    \item Environment Navigation (Lanzi and Loiacono, 2007)
\end{itemize}
\\ \hline

\textbf{Ellipsoidal Based Representation}
& \begin{itemize}
    \item Applicable to real valued environments.
    \item Well-suited in problems wherein the dimensional dependencies are unknown.
\end{itemize}
& \begin{itemize}
    \item More parameters would be evolved in applying high dimensional problems.
    \item Compaction method is needed to reduce the size of evolved rule set. 
    \item Limited by being a particular and symmetric, if orientable, shape.
\end{itemize}
& \begin{itemize}
    \item Function Approximation [Butz, 2005; Butz et al., 2006; Butz et al., 2008]
\end{itemize}
\\ \hline

\textbf{Convex Hull Based Representation}
& \begin{itemize}
    \item Applicable to real valued environments.
    \item Variable length representation.
    \item Fine ability to identify more complex regions.
    \item Generalization of interval and ellipsoidal representation.
    \item Converge faster.
\end{itemize}
& \begin{itemize}
    \item Hard to find set of point to define a convex region.
    \item Hard to define GA operators. 
    \item The number of points needed by convex hulls is exponential with dimensionality.
\end{itemize}
& \begin{itemize}
    \item Function Approximation[Lanzi and Wilson, 2006]
\end{itemize}

\\ \hline

\textbf{Fuzzy Logic Based Representation }
  
(Fuzzy LCS, 
ELF, LFCS, 
FIXCS, 
Fuzzy-XCS, 
Fuzzy-UCS)

& \begin{itemize}
    \item Applicable to real valued environments.
    \item Capability to obtain maximal generalization(representation of the fuzzy rule set as compact as possible)
    \item Fine interpretability of fuzzy rules.
    \item Supporting real valued action.
    \item Robustness to noisy input.
    \item Handling mixed attribute and missing value.
\end{itemize}

& \begin{itemize}
    \item Producing only a limited number of possible rules.
    \item Less freedom in domain knowledge representation and extraction.
    \item Hard to make it flexible by using self-adapting mechanism.
\end{itemize}

& \begin{itemize}
    \item Classification [Valenzuela-Rend$\acute{o}$n, 1991; Orriols-Puig et al., 2007; Orriols-Puig et al., 2008a; Orriols-Puig et al., 2008b]
    \item Robotics [Bonarini, 1993; Bonarini, 1994]
    \item Epidemiologic Classification [Walter and Mohan, 2000]
    \item Reinforcement Problems [Casillas  et al., 2004; Casillas  et al., 2005; Casillas  et al., 2007; Bonarini, 1998]
    \item Function Approximation [Casillas  et al., 2004; Casillas  et al., 2007]
\end{itemize}

\\ \hline
\textbf{First Order Logic Based Representation }
(FOXCS)

& \begin{itemize}
    \item Present the complex relationships among the attributes of a task domain
\end{itemize}

& \begin{itemize}
    \item Hard to define GA operators.
\end{itemize}

& \begin{itemize}
    \item Relational domains [Mellor, 2005]
    \item ILP and RRL tasks [Mellor, 2006; Mellor, 2008]
    \item Classification [Mellor, 2006; Mellor, 2008]
\end{itemize}
\\ \hline

\end{tabular} }
\end{table*}

\begin{table*}[htbp]
{ \scriptsize
\setlength{\tabcolsep}{10pt}
\setlength{\extrarowheight}{2.5pt}
\begin{tabular}{|p{3.5cm}|p{4cm}|p{4cm}|p{4cm}|}
\hline
\textbf{Messy Code Representation} 
(XCSm)

& \begin{itemize}
    \item Variable length representation.
    \item Well-suited in spars problems.
\end{itemize}

& \begin{itemize}
    \item Not suited for real valued problems.
\end{itemize}

& \begin{itemize}
    \item Multi step problems [Lanzi, 1999]
\end{itemize}

\\ \hline

\textbf{GP-like Representation }
  
(XCSL, 
XCSF-GEP, 
DGP-XCS)

& \begin{itemize}
    \item Applicable to real valued and nominal problems.
    \item Able to represent arbitrary regularities.
    \item Offer greater transparency.
    \item Have the ability to ignore unneeded inputs or add ones that become relevant.
\end{itemize}

& \begin{itemize}
    \item Applicable to real valued and nominal problems.
    \item Able to represent arbitrary regularities.
    \item Offer greater transparency.
    \item Have the ability to ignore unneeded inputs or add ones that become relevant.
\end{itemize}

& \begin{itemize}
    \item Function Approximation [Wilson,2008;]
    \item Multi step problems [Preen and Bull, 2009; Lanzi and Perrucci, 1999; , Lanzi, 2003; Cielecki and Unold, 2007]
    \item Data Mining [Lanzi, 2001a]
    \item Learning Context-Free Language [Unold, 2005; Unold and Cielecki, 2005; Unold, 2007]
\end{itemize}

\\ \hline

\textbf{Neural Networks Based Representation}
  
(X-NCS, 
NCS, 
NLCS,
XCSFNN, 
X-NFCS)

& \begin{itemize}
    \item Applicable to real valued environments.
    \item Able to represent arbitrary regularities.
    \item Able to apply on problem with continuous action space.
    \item Able to produce more general classifier.
\end{itemize}

& \begin{itemize}
    \item Low interpretability
    \item Every classifier must accept inputs from all variables, whereas this might not be necessary for some regularities.
\end{itemize}

& \begin{itemize}
    \item Function Approximation [Bull and O'Hara, 2002; Loiacono and Lanzi, 2006]
    \item Multiple Domains [[Bull and O'Hara, 2002; O'Hara and Bull, 2005] 
    \item Robotics [Hurst and Bull, 2004]
    \item Classification [Dam et al., 2008]
\end{itemize}

\\ \hline

\textbf{Tile Coding Based Representation} 
  
(XCSF-RTC)

& \begin{itemize}
    \item Applicable to real valued environments.
    \item XCSF with tile coding prediction converge faster than XCSF with linear approximation.
\end{itemize}

& \begin{itemize}
    \item Hard to evolved the number of tiling and their resolutions.
\end{itemize}

& \begin{itemize}
    \item Reinforcement Problems [Lanzi et al., 2006]
\end{itemize}

\\ \hline

\end{tabular} }
\end{table*} 


\textbf{Problems with mixed attributes:} Hitherto, real valued problems and bit string ones are discussed independently while mixed attributed problems are more common in real word applications. A straightforward solution is considering the problem as two subproblems: one is real valued problem with real valued attribute and the other is nominal problem only containing nominal attributes. These subproblems can be solved individually by two different representation techniques like interval based and messy code representation. Since not all mixed attributed problems can be divided into such subproblems, a representation which can handle both kind of attribute indiscriminately would be useful in such problems. GP-like and fuzzy based representations are some examples of those representation techniques.

In addition to what has already been mentioned, each representation technique has its own remarkable properties which provide an appropriate solution for particular problems. Table 2 summarized the existing knowledge representation approaches, some of their main advantageous, disadvantages, and the problem domain(s) for which the representation was designed and/or tested. We expect that this table to be a roadmap for choosing the best technique among wide varieties of existing knowledge representation approaches proposed so far.

\section{Summary and Conclusion}
\label{sec:12}
In this study, we focus on the progress of XCS as a rule based learning system with a special consideration into the improvement of one of its most important components known as knowledge representation. As this component seeks prominent role in generalization capacity of the system and impacts on rule based system in terms of efficiency and efficacy, the past decade has seen a growing interest in proposing different knowledge representation techniques in LCS domain in general and XCS in specific. Here, after reviewing some basic information, we survey different knowledge representation techniques proposed in Michigan LCS realm and grouping them into different categories based on the classification approach in which they are incorporated. In each category, the underlying rule representation schema and the format of classifier condition to support the corresponding representation are presented. Furthermore, a precise explanation on the way that each technique partitions the problem space along with the extensive experimental results is provided. The main properties of knowledge representation technique, investigated in this paper, can be used as an illumination guideline to determine an appropriate knowledge representation technique for the domain in question. To shed light on these properties, a comparative analysis on some conventional problems is provided in general perspective. 

We hope that current review facilitates a better understanding of the different streams of research on this topic and underlies proper usage of LCS in many applications. In addition, since knowledge representation is a typical component of any rule based system like XCS, the current research can be a roadmap in other rule based systems to represent the problem regularities properly. Furthermore, we anticipate this survey to be interest to the LCS researchers and practitioners since it provides a guideline for choosing a proper knowledge representation technique for a given problem and also opens up new streams of research on this topic.


\bibliographystyle{spbasic}      


\end{document}